\documentclass[journal]{IEEEtranTIE}
\usepackage{graphicx}
\usepackage{cite}
\usepackage{picinpar}
\usepackage{amsmath}
\usepackage{url}
\usepackage{flushend}
\usepackage[utf8]{inputenc}
\usepackage{colortbl}
\usepackage{soul}
\usepackage{multirow}
\usepackage{pifont}
\usepackage{color}
\usepackage{alltt}
\usepackage[hidelinks]{hyperref}
\usepackage{enumerate}
\usepackage{siunitx}
\usepackage{breakurl}
\usepackage{epstopdf}
\usepackage{pbox}
\usepackage{newtxmath} 
\usepackage{times} 
\usepackage{amsmath}

\usepackage{amssymb}
\usepackage[linesnumbered,ruled,lined]{algorithm2e}
\usepackage{subfigure}
\usepackage{textcomp}
\usepackage[dvipsnames]{xcolor}
\usepackage{booktabs}

\begin{document}
\title{	Adaptive Model Prediction Control-Based Multi-Terrain Trajectory Tracking Framework for Mobile Spherical Robots}

\author{
	\vskip 1em
	
	Yifan Liu, \emph{Student Member}, \emph{IEEE}, Tao Hu, Xiaoqing Guan, Yixu Wang, Bixuan Zhang,\\You Wang, \emph{Member}, \emph{IEEE}, Guang Li, \emph{Member}, \emph{IEEE}

	\thanks{
	
		This work was supported by the Fundamental Research Funds for the Central Universities 226-2022-00086. (Corresponding author: You Wang.)
		
		Yifan Liu, Tao Hu, Xiaoqing Guan, Yixu Wang, Bixuan Zhang, You Wang and Guang Li are with State Key Laboratory of Industrial Control Technology, Institute of Cyber Systems and Control, Zhejiang University, Hangzhou, 310027, China. (e-mail: \{yifanliu, hutao, xiaoqing\_guan, yixuwang, bixuan\_zhang, king\_wy, guangli\}@zju.edu.cn). 
	}
}

\maketitle

\begin{abstract}
Owing to uncertainties in both kinematics and dynamics, the current trajectory tracking framework for mobile robots like spherical robots cannot function effectively on multiple terrains, especially uneven and unknown ones. Since this is a prerequisite for robots to execute tasks in the wild, we enhance our previous hierarchical trajectory tracking framework to handle this issue. First, a modified adaptive RBF neural network (RBFNN) is proposed to represent all uncertainties in kinodynamics. Then the Lyapunov function is utilized to design its adaptive law, and a variable step-size algorithm is employed in the weights update procedure to accelerate convergence and improve stability. Hence, a new adaptive model prediction control-based instruction planner (VAN-MPC) is proposed. Without modifying the bottom controllers, we finally develop the multi-terrain trajectory tracking framework by employing the new instruction planner VAN-MPC. The practical experiments demonstrate its effectiveness and robustness. 
\end{abstract}



\definecolor{limegreen}{rgb}{0.2, 0.8, 0.2}
\definecolor{forestgreen}{rgb}{0.13, 0.55, 0.13}
\definecolor{greenhtml}{rgb}{0.0, 0.5, 0.0}

\section{INTRODUCTION}
\IEEEPARstart{T}{he} spherical robot is a special type of mobile robot that has potential applications in a variety of fields, including environmental detection, search and rescue, and security patrols. Due to its unique mechanical structure, the robot has two major advantages over conventional mobile robots such as wheeled robots and legged robots \cite{bib3, bib10, pre2}. First, its closed spherical shell can assist in protecting the interior electrical and mechanical framework from collision and damage. Second, it has flexible motion and low frictional energy consumption due to its spherical rolling mechanism. However, despite these benefits, the spherical robot faces many challenges in the field of control due to its non-linear and under-actuated characteristics\cite{bib17}.

Trajectory tracking control plays an important role for mobile robots since it is the foundation for perfectly implementing the task and plan. In practical applications, many tasks, such as environmental detection, require robots that can operate effectively on multiple terrains. Unfortunately, the lack of an efficient multi-terrain trajectory tracking algorithm has prevented spherical robots from performing such tasks. We believe the problem with control on multiple terrains is due to the uncertainties in the models, including the kinematics and dynamics. There are two possible causes for these uncertainties, i.e., unmodeled structures in the kinodynamic model such as unknown disturbances from the environment, and uncertainties in parameters like friction. In addition, these uncertainties may vary on different terrains \cite{pre2, adap1}. 

\begin{figure}[tb]
\centering
\includegraphics[width=8cm]{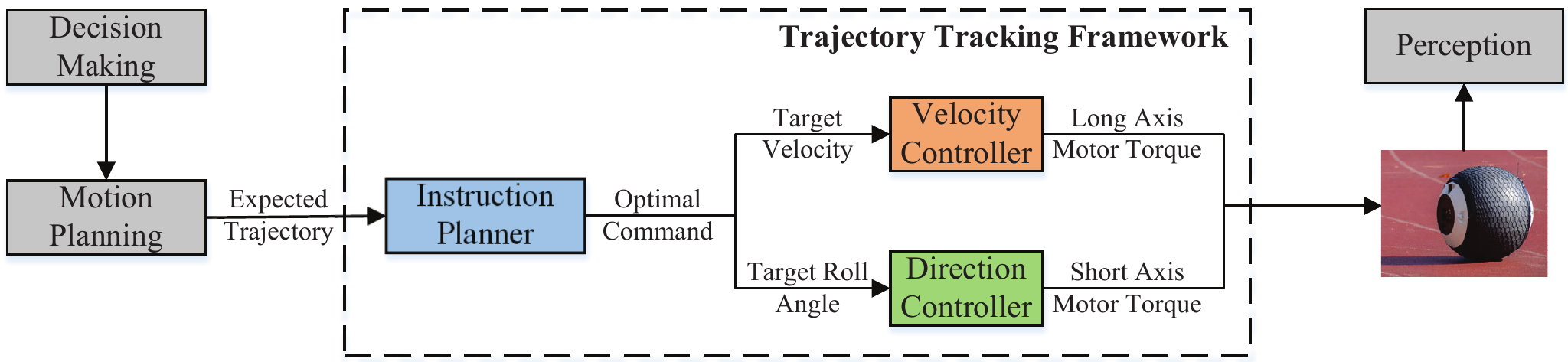}
\caption{Trajectory Tracking Framework of the spherical robot.}
\label{fig1}
\vspace{-0.5cm}
\end{figure}

In recent years, numerous works on the control of spherical robots have been published. Most research concentrates on the velocity control with the PID controller and sliding mode controller (SMC) \cite{bib10}. Besides, \cite{bib29} studies the motion planning of spherical robots for obstacle avoidance problems. There are other studies on trajectory tracking and instruction planning \cite{bib33}. However, they only provide simulation results. Then an efficient velocity controller called hierarchical SMC (HSMC) and an effective direction controller called hierarchical terminal SMC (HTSMC) are proposed by us \cite{pre4}. With an MPC-based instruction planner, we establish the trajectory tracking framework MHH (MPC-HSMC-HTSMC), which is the state of the art (SOTA) \cite{pre4}. Fig.~\ref{fig1} illustrates the framework. However, this framework can only be used on certain flat tiled terrain without considering uncertainties.

Uncertainty problems and multi-terrain control are important research fields in robotics' control, such as autonomous ground vehicles (AGV), manipulators, and surface vehicles \cite{adap2, rl5}. \cite{rl5} proposes the model-reference reinforcement learning control method for autonomous surface vehicles and obtains good results in simulation despite requiring prior training. \cite{rl1} presents a reinforcement learning (RL) method to train a trajectory tracking controller for quadrupedal robots without modifying the bottom whole-body controller. For the multi-terrain control of quadrupedal robotics, \cite{rl3, rl4} use RL to train the robots on different terrains. However, for the quadrupedal robot, its foot sensor can detect terrain quite effectively, whereas the spherical robot has no such advantages \cite{terrain1}. Moreover, model-based RL has a close relationship with the uncertainty problem \cite{rl6, rl2, rl9}. Though this method may not be fast enough and cannot ensure stability, we can combine this structure with a Lyapunov-based adaptive law.

As for adaptive algorithms, some studies employ radial basis function neural network (RBFNN) due to its advantages of fast convergence speed and strong mapping ability \cite{bib26}. In our previous work \cite{pre2}, a multi-terrain velocity controller is proposed by integrating an RBFNN and the HSMC. However, it only involves half of the dynamics, and there is only one uncertainty to be predicted. In addition, traditional RBFNN usually has one output, which cannot be directly employed in the trajectory tracking problem. In terms of AGV, \cite{adap2, adap4} design neural network or Kalman filter to estimate the uncertainties in a single-level system. \cite{adap3} combines their previous control Lyapunov function-MPC (CLF-MPC) with a uncertainties estimation matrix. The problem is that this method only suits the dynamic-model-based controller as its pivotal regressor and skew-symmetry matrix come from the significant properties of the dynamic model that is proposed in \cite{bib39}. 

Online learning is another kind of method to obtain the uncertainties. In \cite{online1}, they employ a Gaussian process regression to estimate the uncertainties in the manipulator's dynamics, and integrate the torque compensation with the feedback linearization controller. But it may not be effective for multi-terrain problems, as the accumulated data sets will generate erroneous values when terrain changes. And it cannot manage variable dynamics or disturbances. Moreover, some online learning algorithms, such as FTRL (Follow The Regularized Leader) and the SGD (Stochastic Gradient Descent) family, provide us with fresh ideas, namely that the variable step-size algorithm can be employed to the adaptive algorithm, thereby enhancing the learning speed and stability \cite{bib40, bib41}.

In this work, we structure the uncertainties in both dynamics and kinematics in the form of command compensations and then insert the uncertainties into MPC's kinematic model. A multi-output modified RBFNN is employed to estimate them and update their weights using a Lyapunov-based adaptive law. A variable step-size algorithm designed for the adaptive algorithm of neural networks is also applied to the weight update procedure, thereby accelerating convergence and enhancing stability. And in this way, we obtain the adaptive instruction planner VAN-MPC (Adaptive RBFNN CLF-MPC with Variable Step-Size). In one step, an efficient multi-terrain trajectory tracking control framework VANMHH (VAN-MPC-HSMC-HTSMC) is provided by replacing the original instruction planner MPC with VAN-MPC, without modifying the dynamic-based bottom controllers. Experiments on multiple terrains also confirmed its efficiency. Specifically, this study makes the following contributions:
\begin{itemize}
    \item The first multi-terrain trajectory tracking control framework is proposed for spherical robots. And an efficient adaptive MPC instruction planner is presented.
    
    \item By optimizing the instruction planner only, multi-terrain task is accomplished without further changes to the bottom controllers. This offers great portability and convenience and can be applied to other mobile robots' uncertainty problem. After sorting, relevant code will be open-sourced and posted to https://github.com/Ivanniour/Adaptive-MPC-for-Robots-Multi-Terrain-Trajectory-Tracking.

    \item The variable step-size algorithm designed for the adaptive controller can accelerate convergence and enhance stability, and can be applied to other adaptive algorithms.

    \item Uncertainties in the whole system are structured and estimated with a multi-output modified RBFNN in a MPC-based planner.
\end{itemize}

The paper is organized as follows: Section \uppercase\expandafter{\romannumeral2} discusses the uncertain problem and associated background. Section \uppercase\expandafter{\romannumeral3} presents the adaptive hierarchical trajectory tracking algorithm for multi-terrain tracking control. In section \uppercase\expandafter{\romannumeral4}, practical experiments are carried out, results and comparisons are given. Finally, conclusions are proposed in Section \uppercase\expandafter{\romannumeral5}.

\section{BACKGROUND AND PROBLEM FORMULATION}
\begin{figure}[tb]
\centering
\subfigure[Schematic illustration]
{
    \label{fig2:subfig:a} 
    \includegraphics[width=3cm]{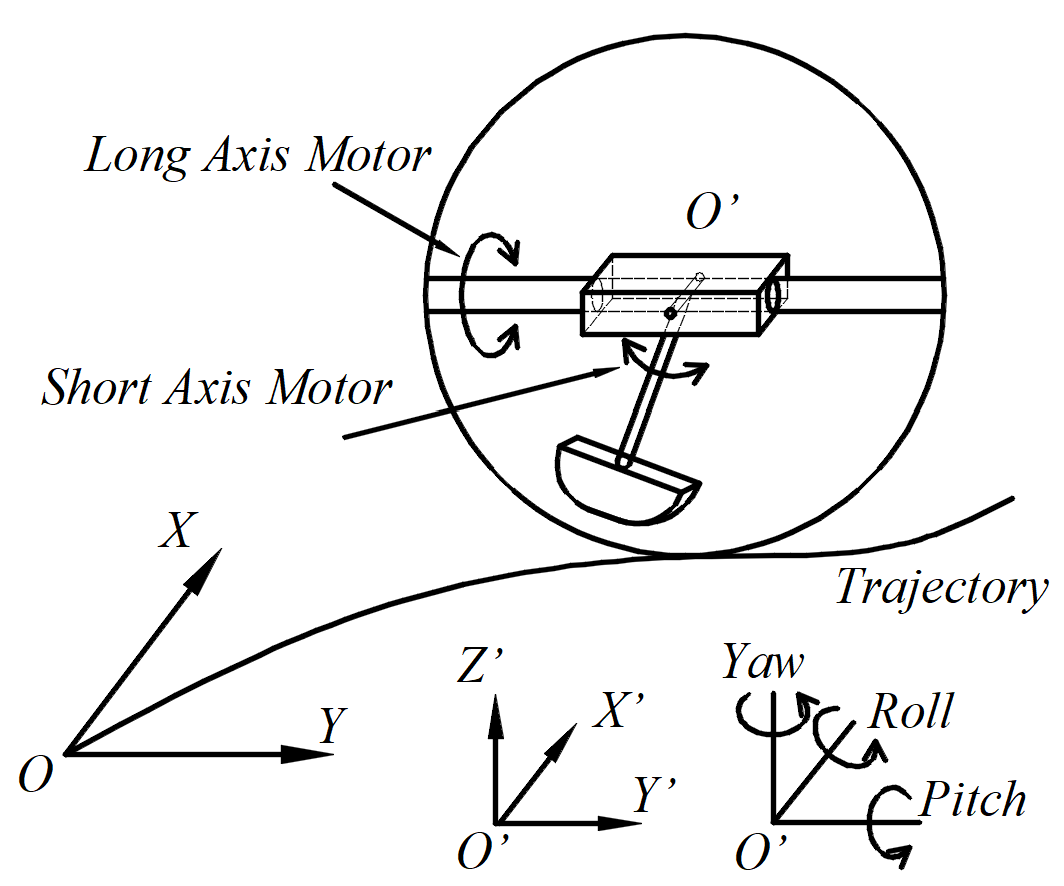}}
\subfigure[Mechanical structure]
{
    \label{fig2:subfig:b} 
    \includegraphics[width=3cm]{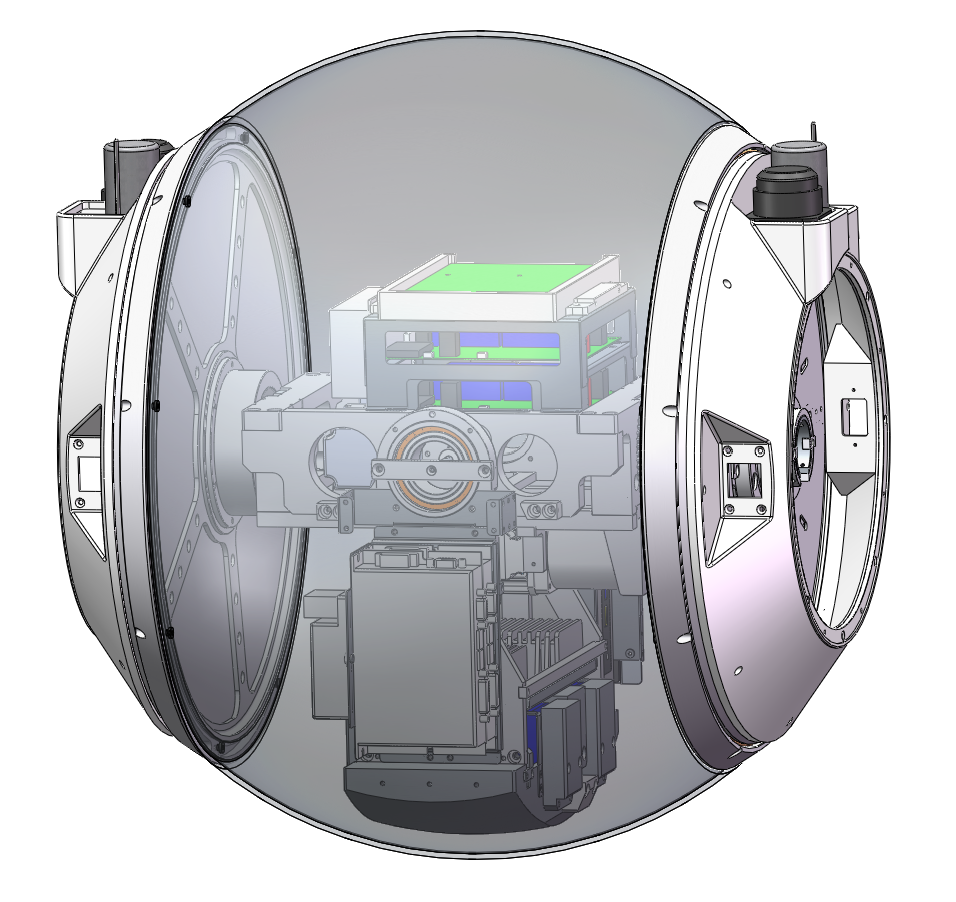}}
    \hspace{0in}
\caption{Mechanical schematic of the spherical robot. The frame OXY is the world frame where O is the frame origin. And O'X'Y'Z' is the local frame centered on the robot O', with X' representing the forward direction \cite{pre4}.}
\vspace{-0.3cm}
\label{fig2}
\end{figure}

The spherical robot is driven by a 2-DOF heavy pendulum and the mechanical schematic of the robot can be seen in Fig.~\ref{fig2}, and two motors are employed to achieve the robot's locomotion. In our previous work, we have proposed an efficient hierarchical trajectory tracking framework MHH for certain flat tiled floor \cite{pre4}. However, on different terrains, especially uneven terrains, MHH's control effect will be drastically decreased due to the uncertainties. In this section, we will first introduce our previous framework MHH. Then, the problem will be clarified and the uncertainties will be structured in subsection B. 

\subsection{Hierarchical Trajectory Tracking Framework}
\begin{figure}[tb]
\centering
\includegraphics[width=6cm]{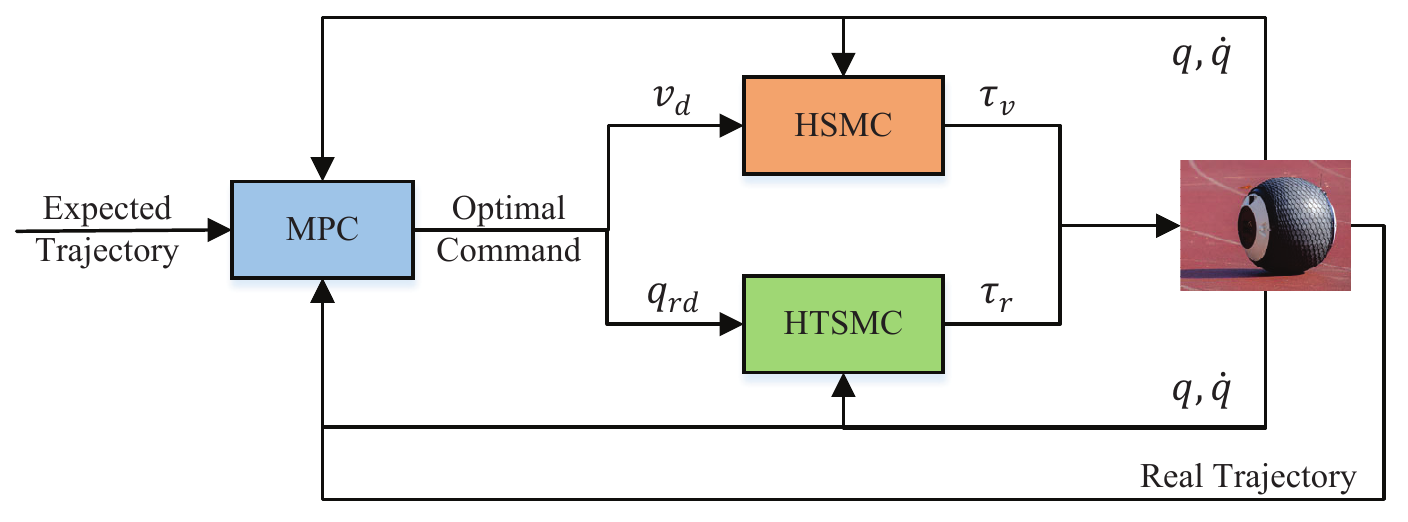}
\caption{The control scheme of MHH.}
\label{fig3}
\end{figure}

A hierarchical trajectory tracking control framework MHH is developed by us for spherical robots in \cite{pre4}, and Fig.~\ref{fig3} illustrates its framework. An MPC controller is designed to find the optimal command based on the reference trajectory and the estimated kinematic model. Define $\boldsymbol{x}=\begin{bmatrix} X & Y & \phi \end{bmatrix}^T$ and $\boldsymbol{u}=\begin{bmatrix} v & q_r \end{bmatrix}^T$ are the states and inputs of the system. $X$, $Y$ represent the coordinates in the global frame OXY in Fig. \ref{fig1}. $\phi$, $v$ and $q_r$ represent the robot's yaw angle, forward velocity, and roll angle, respectively. $R$ is the radius of spherical robot. The estimated kinematic model can be seen as follows in \eqref{eq2}.

\begin{equation}
\begin{bmatrix}
    \dot{X} \\ \dot{Y} \\ \dot{\phi}
\end{bmatrix}
=
\boldsymbol{\hat{f}}\left(\boldsymbol{x}, \boldsymbol{u}\right)=
\begin{bmatrix}
    v\cos\phi\\
    v\sin\phi\\
    v\tan q_r/R
\end{bmatrix}
\label{eq2}
\end{equation}

After optimal commands $v$ and $q_r$ are determined from the MPC, two model-based torque controllers execute the command based on the dynamic model and control laws. The estimated dynamic model is represented by \eqref{eq3}, which is derived from the Euler-Lagrange equations \cite{pre4}.
\begin{equation}
\boldsymbol{\hat{M}}\left(\boldsymbol{q}\right)\boldsymbol{\ddot{q}}+\boldsymbol{\hat{N}}\left(\boldsymbol{q},\boldsymbol{\dot{q}}\right)=\boldsymbol{E}\boldsymbol{{\hat\tau}}
\label{eq3}
\end{equation}
where $\boldsymbol{\hat{M}}\left(\boldsymbol{q}\right)\in{\mathbb{R}^{4\times4}}$ and $\boldsymbol{\hat{N}}\left(\boldsymbol{q},\boldsymbol{\dot{q}}\right)\in{\mathbb{R}^{4}}$ is the estimated inertia matrix and nonlinear matrix, respectively. $\boldsymbol{q}$ is the state matrix in \cite{pre4} and $\boldsymbol{\hat\tau}=\begin{bmatrix}
\hat{\tau}_v & \hat{\tau}_r\end{bmatrix}^T$ represents its input torque matrix.

\begin{figure}[tb]
\centering
\subfigure[HSMC' sliding surface]
{
    \label{fig4:subfig:a} 
    \includegraphics[width=4.25cm]{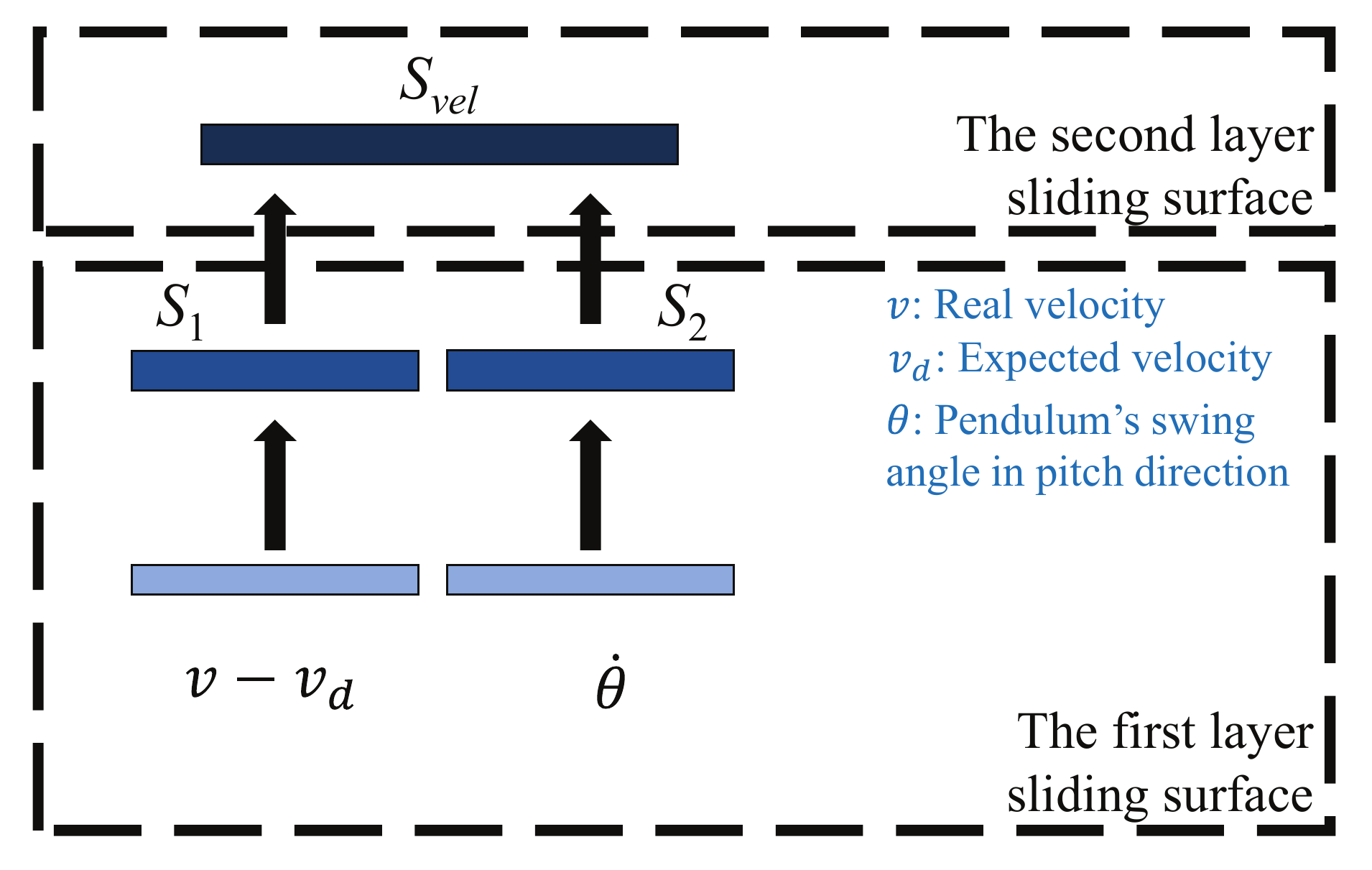}}
    \hspace{-0.1in}    
\subfigure[HTSMC' sliding surface]
{
    \label{fig4:subfig:b} 
    \includegraphics[width=4.25cm]{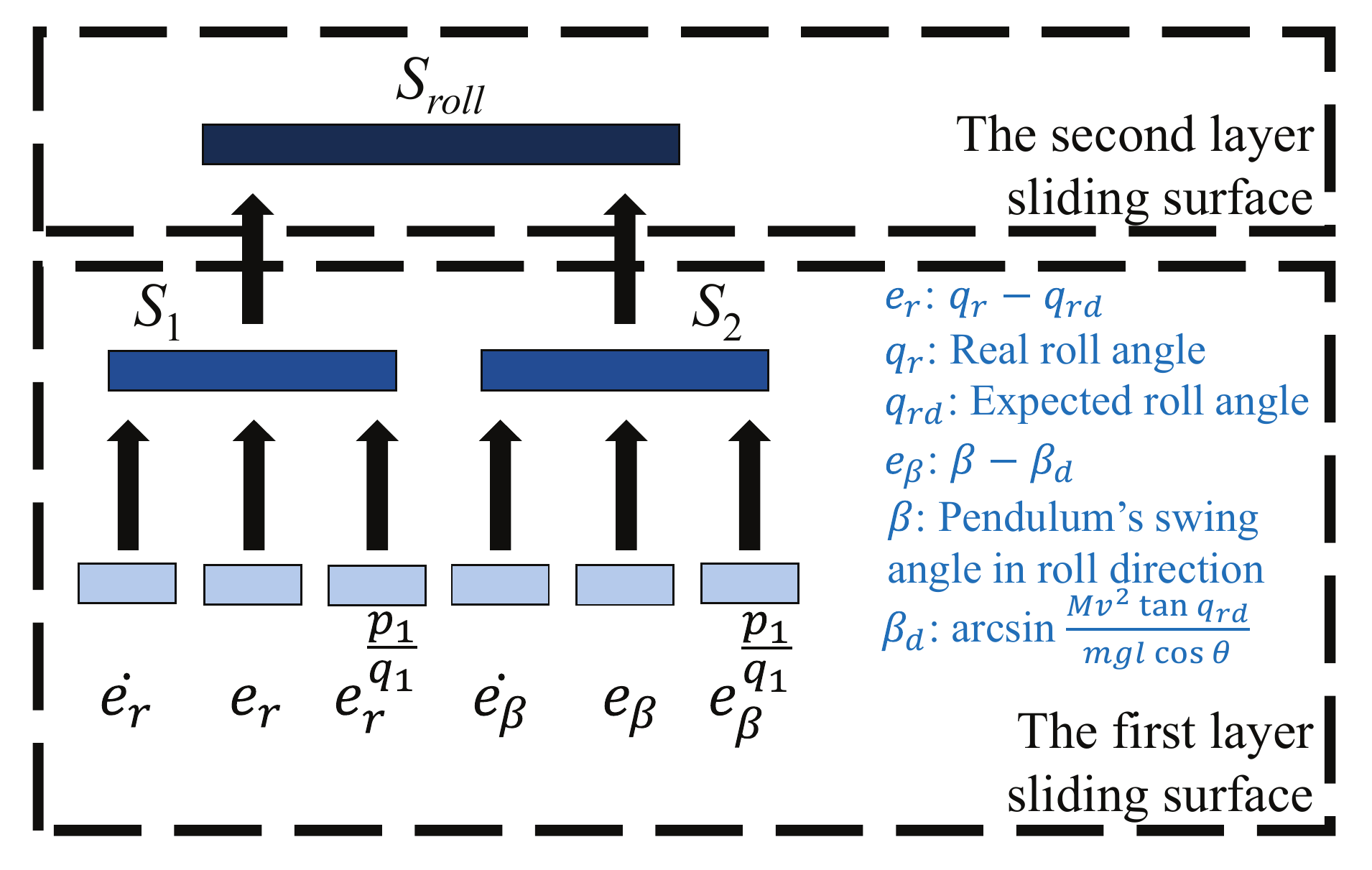}}
    \hspace{0in}
\caption{Sliding surfaces of the two hierarchical SMCs. HSMC is the velocity controller while HTSMC is applied to control the roll angle and direction.}
\label{fig4}
\end{figure}

The velocity and direction controllers above are all hierarchical SMCs, whose sliding surfaces are composed of both a control surface and a stability surface, as can be seen in Fig.~\ref{fig4}. These two controllers' control laws can be abbreviated as $F_{vel}$ and $F_{roll}$ for convenience. What's more, both of the two controllers are dynamic model-based controllers, and we can get \cite{pre4}:
\begin{equation}
\begin{bmatrix}
    \hat{\tau}_v \\ \hat{\tau}_r
\end{bmatrix}
= 
\begin{bmatrix}
    F_{vel}\left(\boldsymbol{q}, \boldsymbol{\dot{q}}, v_{d} \right)\\
    F_{roll}\left(\boldsymbol{q}, \boldsymbol{\dot{q}}, q_{rd} \right)
\end{bmatrix}
\label{eq4}
\end{equation}
where $v_d$ and $q_{rd}$ represent, respectively, command velocity and roll angle (which determines direction) from MPC.

\subsection{Problem Formulation}
Both the estimated kinematic model in \eqref{eq2} and the estimated dynamic model in \eqref{eq3} are theoretical models derived from flat tiled terrain. However, when it moves on different terrains, the previous MHH may perform poorly due to the uncertainties between the real and estimated kinodynamics. We intend to express the uncertainties and predict them online, so as to accomplish multi-terrain trajectory tracking control \cite{pre2}. 

The problem, however, is that if we estimate the uncertainties of all three controllers simultaneously, the entire control framework would become bloated. Furthermore, since these three components of uncertainties operate independently, they may contradict one another. Therefore, we intend to design a one-step solution and incorporate all uncertainties into the kinematics, so that modifying the instruction planner MPC is all that is necessary to achieve the purpose. The detailed derivation is described below.

The first step is to transform the uncertainties in the dynamics into command compensation. Take the following equation as the real dynamic model of the spherical robot.
\begin{equation}
\boldsymbol{M}\left(\boldsymbol{q}\right)\boldsymbol{\ddot{q}}+\boldsymbol{N}\left(\boldsymbol{q},\boldsymbol{\dot{q}}\right)=\boldsymbol{E}\boldsymbol{\tau}
\label{eq11}
\end{equation}
where $\boldsymbol{M}\in{\mathbb{R}^{4\times4}}$ and $\boldsymbol{N}\in{\mathbb{R}^{4\times4}}$. According to \cite{pre2}, we can denote all uncertainties with $\Delta\boldsymbol{\tau}$ after derivation, then:

\begin{equation}
\boldsymbol{E}^{-1}\left[\boldsymbol{M}\left(\boldsymbol{q}\right)\boldsymbol{\ddot{q}}+\boldsymbol{N}\left(\boldsymbol{q},\boldsymbol{\dot{q}}\right)\right]=\boldsymbol{\hat\tau} +\boldsymbol{\Delta \tau}
\label{eq12}
\end{equation}

During the trajectory tracking control process, the input torque $\boldsymbol{\hat\tau}$ is obtained from the two SMCs' control laws in \eqref{eq4}. And it should be emphasized that the input would solely depends on the commands $v_d$ and $q_{rd}$ at the moment when the states $\left( \boldsymbol{q}, \;\; \boldsymbol{\dot{q}}\right)$ are determined, according to \cite{pre4}. Therefore, we can make the following derivation:

\begin{equation}
\begin{aligned}
\boldsymbol{E}^{-1}\left[\boldsymbol{M}\left(\boldsymbol{q}\right)\boldsymbol{\ddot{q}}+\boldsymbol{N}\left(\boldsymbol{q},\boldsymbol{\dot{q}}\right)\right]=&
\begin{bmatrix}
    F_{vel}\left(\boldsymbol{q}, \boldsymbol{\dot{q}}, v_{d} \right)\\
    F_{roll}\left(\boldsymbol{q}, \boldsymbol{\dot{q}}, q_{rd} \right)
\end{bmatrix} +
\begin{bmatrix}
    \Delta\tau_v \\
    \Delta\tau_r
\end{bmatrix} \\
=&\begin{bmatrix}
    F_{vel}\left(\boldsymbol{q}, \boldsymbol{\dot{q}}, v_{d}+\Delta v_{d}^D \right)\\
    F_{roll}\left(\boldsymbol{q}, \boldsymbol{\dot{q}}, q_{rd}+\Delta q_{rd}^D \right)
\end{bmatrix} 
\label{eq13}
\end{aligned}
\end{equation}

Define $\boldsymbol{\Delta u_{D}} = \begin{bmatrix}\Delta v_{d}^D & \Delta q_{rd}^D\end{bmatrix}^T$ to represent the uncertainties in the whole-body dynamic model. 
Then, transfer $\boldsymbol{\Delta u_{D}}$ to the kinematics, and the real kinematic model $\boldsymbol{f}\left(\boldsymbol\cdot\right)$ will become:

\begin{equation}
\boldsymbol{\dot{x}}=\boldsymbol{f}\left(\boldsymbol{x}, \boldsymbol{u}\right)=
\boldsymbol{\hat{f}}\left(\boldsymbol{x}, \boldsymbol{u}-\boldsymbol{\Delta u_{D}}\right)-\boldsymbol{\Delta\delta}
\label{eq14}
\end{equation}
where $\boldsymbol{\Delta\delta}\in{\mathbb{R}^{3}}$ is the original uncertainties in the kinematic model.

At present, all of the uncertainties have been included in the kinematic model. However, the existing phrase is too complex and lacks sufficient intuitiveness. In addition, $\boldsymbol{\Delta u_{D}}$ is inside the nonlinear model $\boldsymbol{\hat f}\left(\cdot\right)$, which makes it challenging to cope with. Then we derive the following:

\begin{equation}
\begin{aligned}
\boldsymbol{\dot{x}}=&\boldsymbol{f}\left(\boldsymbol{x}, \boldsymbol{u}\right)=
\boldsymbol{\hat{f}}\left(\boldsymbol{x}, \boldsymbol{u}-\boldsymbol{\Delta u_{D}}\right)-\boldsymbol{\Delta\delta}\\
\approx&\boldsymbol{\hat{f}}\left(\boldsymbol{x_0}, \boldsymbol{u_0}\right) + \boldsymbol{\hat{f}_{x}^{'}} \times\left(\boldsymbol{x}-\boldsymbol{x_0}\right) + \boldsymbol{\hat{f}_{u}^{'}}\times\left(\boldsymbol{u}-\boldsymbol{u_0}-\boldsymbol{\Delta u_{D}}\right)-\boldsymbol{\Delta\delta}
\label{eq15}
\end{aligned}
\end{equation}
where we abbreviate $\frac{\partial \boldsymbol{\hat f}}{\partial \boldsymbol{x}}\Big|_{\boldsymbol{x_0},\boldsymbol{u_0}}$ and $\frac{\partial \boldsymbol{\hat f}}{\partial \boldsymbol{u}}\Big|_{\boldsymbol{x_0},\boldsymbol{u_0}}$ as $\boldsymbol{\hat{f}_{x}^{'}}$ and $\boldsymbol{\hat{f}_{u}^{'}}$, respectively. Define $\boldsymbol{\Delta u_K} = \boldsymbol{\hat{f}_{u}^{'-1}}\boldsymbol{\Delta\delta}$, then we get:
\begin{equation}
\begin{aligned}
\boldsymbol{\dot{x}}\approx&\boldsymbol{\hat{f}}\left(\boldsymbol{x_0}, \boldsymbol{u_0}\right) + \boldsymbol{\hat{f}_{x}^{'}}\times\left(\boldsymbol{x}-\boldsymbol{x_0}\right) + \boldsymbol{\hat{f}_{u}^{'}}\times\left(\boldsymbol{u}-\boldsymbol{u_0}-\boldsymbol{\Delta u_{D}}-\boldsymbol{\Delta u_K}\right)\\
\approx&\boldsymbol{\hat f}\left(\boldsymbol{x}, \boldsymbol{u}\right)-\boldsymbol{\hat{f}_{u}^{'}}\times\left(\boldsymbol{\Delta u_{D}}+\boldsymbol{\Delta u_K}\right)
\label{eq16}
\end{aligned}
\end{equation}

Define $\boldsymbol{\Delta u_{D}}+\boldsymbol{\Delta u_K}=\boldsymbol{\Delta u}\in{\mathbb{R}^{2}}$, then the final real kinematic model is shown as follows:
\begin{equation}
\boldsymbol{\dot{x}}=\boldsymbol{f}\left(\boldsymbol{x}, \boldsymbol{u}, \boldsymbol{\Delta u}\right)=\boldsymbol{\hat f}\left(\boldsymbol{x}, \boldsymbol{u}\right)-\boldsymbol{\hat{f}_{u}^{'}} \boldsymbol{\Delta u}
\label{eq17}
\end{equation}

$\boldsymbol{\hat{f}_{u}^{'}} \boldsymbol{\Delta u}$ is the term representing all of the uncertainties in the kinodynamics. And $\boldsymbol{\Delta u}=\begin{bmatrix}
    {\Delta v} & {\Delta q_r}
\end{bmatrix}^T$ can be seen as the input compensations employed to eliminate the aforementioned uncertainties. Though $\boldsymbol{\Delta u}$ is actually unknown, we can use one neural network to represent it and update its value online to approach the real value.

\begin{figure}[tb]
\centering
\includegraphics[width=7cm]{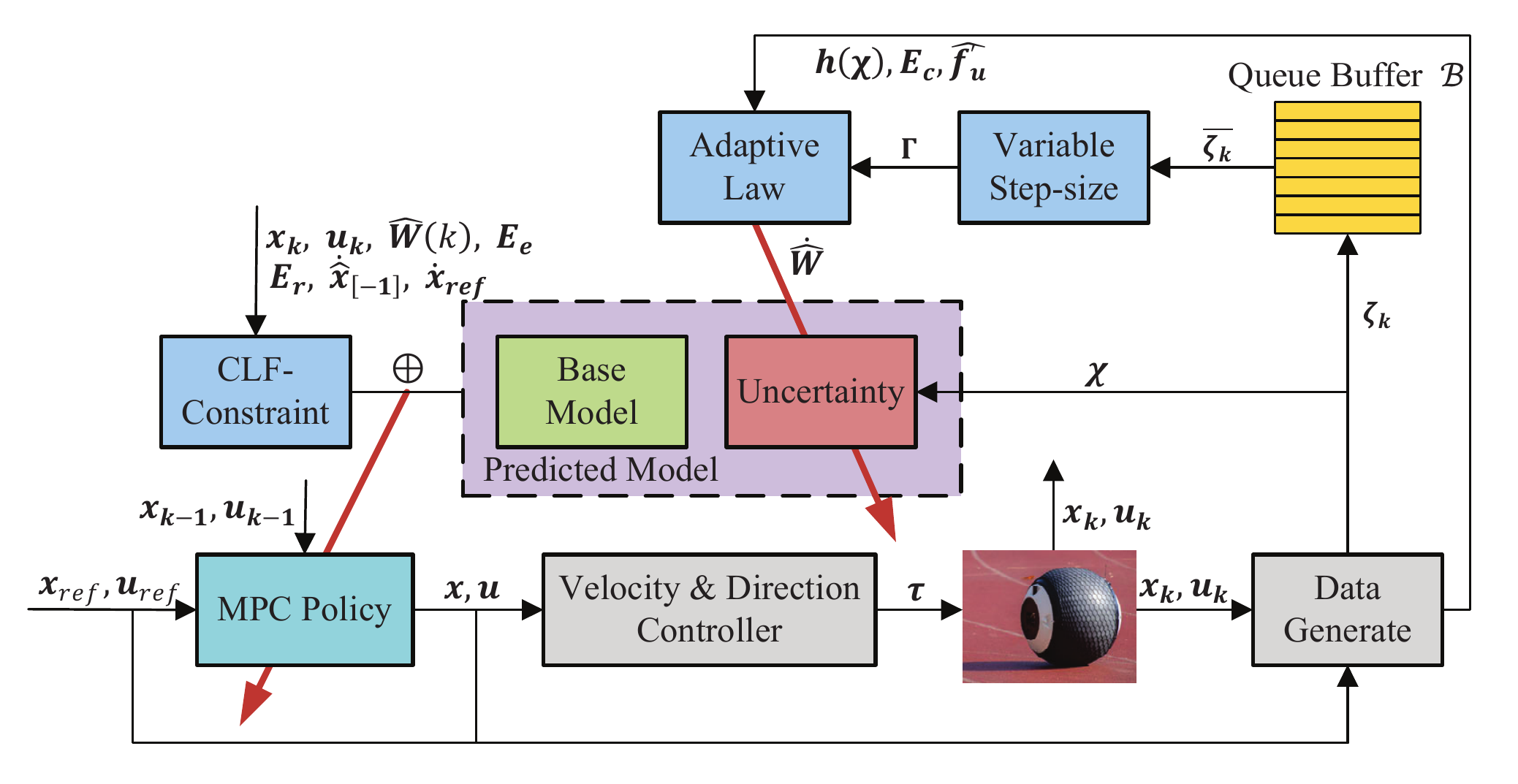}
\caption{The control scheme of VAN-MPC. Variables are described in the following sections.}
\label{fig5}
\vspace{-0.3cm}
\end{figure}
\begin{figure}[tb]
\centering
\includegraphics[width=7cm]{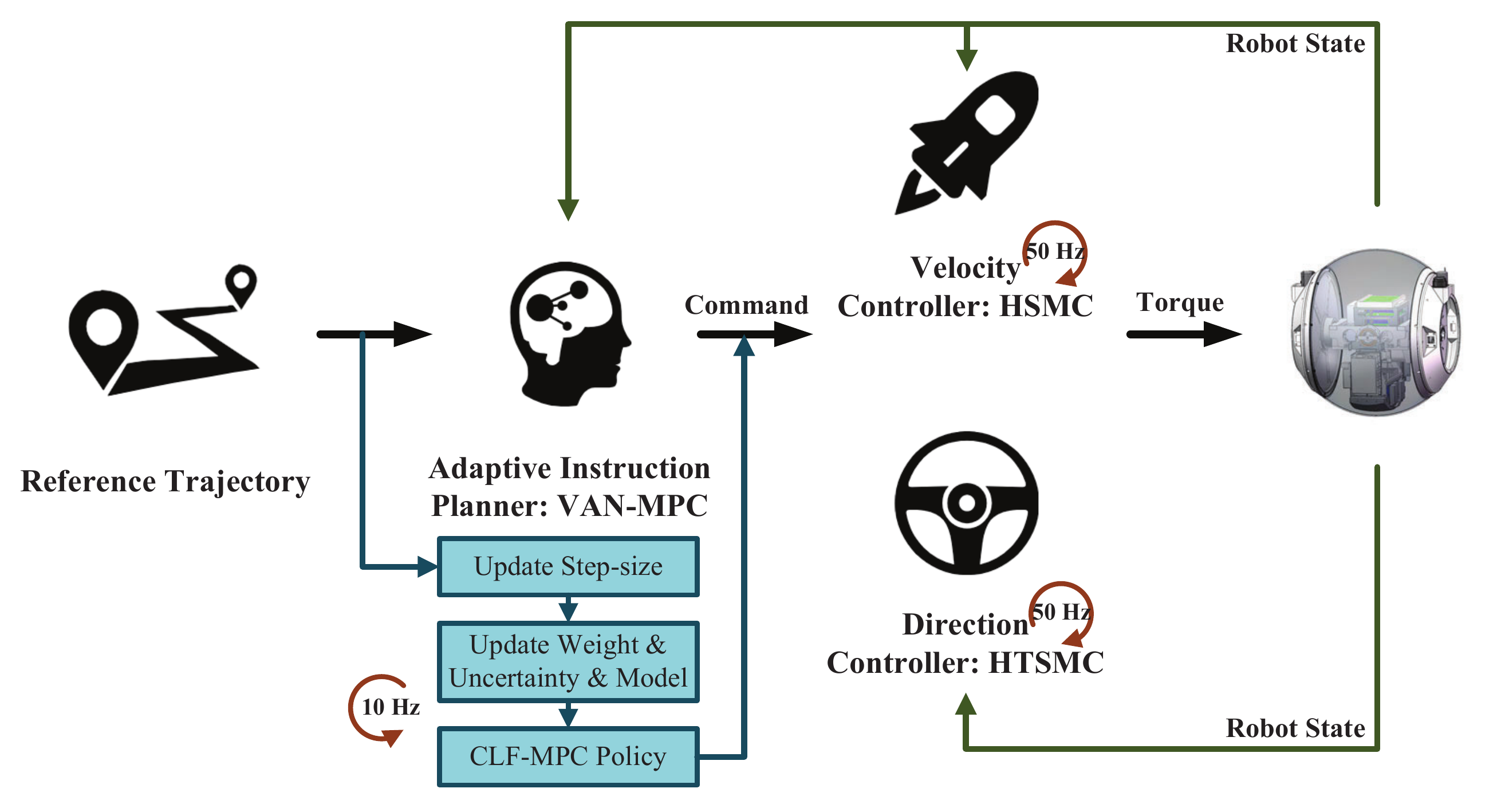}
\caption{Multi-terrain trajectory tracking framework VANMHH.}
\label{fig6}
\vspace{-0.3cm}
\end{figure}

\section{CONTROL ALGORITHM}
In this section, first, a new modified RBF neural network is employed to represent all of the uncertainties in subsection \uppercase\expandafter{\romannumeral3}.A. To enable multi-terrain trajectory tracking, we then propose the adaptive RBF neural network CLF-MPC (AN-MPC) to replace the previous instruction planner MPC in subsection \uppercase\expandafter{\romannumeral3}.B. Furthermore, a corresponding variable step-size algorithm is provided for the adaptive law to accelerate the adaptation process and maintain stability. With the above variable step-size algorithm, a more efficient instruction planner VAN-MPC is developed in subsection \uppercase\expandafter{\romannumeral3}.C. With VAN-MPC, the previous HSMC and HTSMC, we finally propose the multi-terrain trajectory tracking framework VANMHH. The schemes of the VAN-MPC and VANMHH can be seen in Fig.~\ref{fig5} and Fig.~\ref{fig6}, respectively. Details about the algorithm are provided below.

\subsection{Uncertainties and RBFNN}
\begin{figure}[tb]
\centering
\includegraphics[width=6cm]{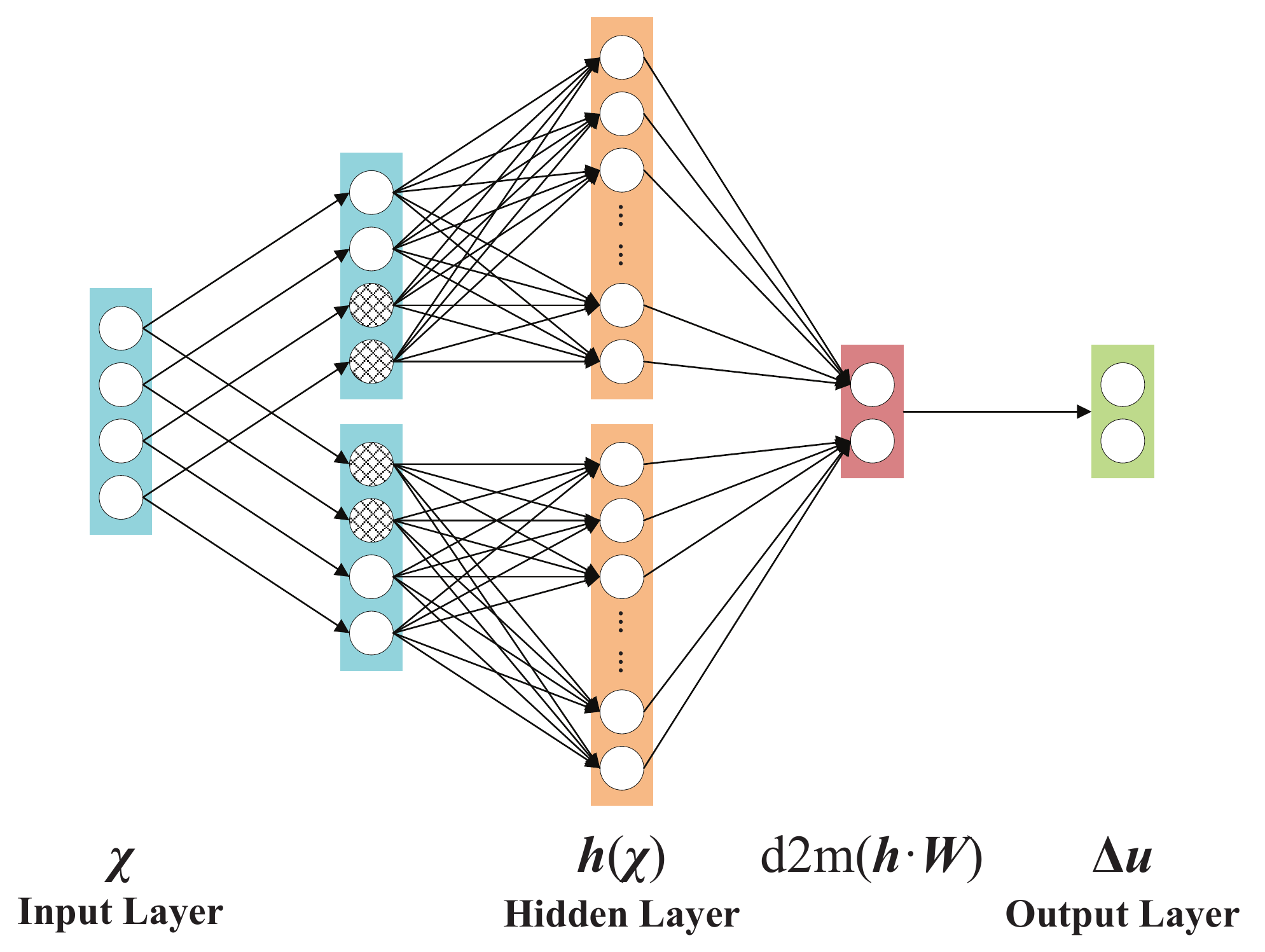}
\caption{The structure of our modified RBFNN.}
\label{fig7}
\vspace{-0.4cm}
\end{figure}

According to \eqref{eq17}, the unknown term $\boldsymbol{\Delta u}=\begin{bmatrix}
    \Delta v & \Delta q_r
\end{bmatrix}^T$ refers to the command compensation utilized to eliminate the uncertainties in the whole system. To fit the unknown term $\boldsymbol{\Delta u}$, we design a special multiple-input and multiple-output RBFNN as follows. The structure of the modified RBFNN is shown in Fig.~\ref{fig7}.

\begin{equation}
\begin{aligned}
\boldsymbol{\Delta u}
&=\boldsymbol{\Gamma}\text{d2m}\left(\boldsymbol{h}\left(\boldsymbol{\chi}\right)\boldsymbol{W}_*\right)=\boldsymbol{\Gamma} \begin{bmatrix}
\boldsymbol{h_1 W_{1}} \\ \boldsymbol{h_2 W_{2}}
\end{bmatrix}
\label{eq18}
\end{aligned}
\end{equation}

\begin{equation}
\text{d2m}(\boldsymbol A) = \begin{bmatrix} 1&0\\0&0\end{bmatrix} A \begin{bmatrix} 1\\0\end{bmatrix} + \begin{bmatrix} 0&0\\0&1\end{bmatrix} A \begin{bmatrix} 0\\1\end{bmatrix}
\label{eq19}
\end{equation}

\begin{equation}
h_{kj}\left(\boldsymbol{\chi}\right)=\exp{\left(-\frac{\Vert\boldsymbol{\chi}-\boldsymbol{c_{j}}\Vert^2_{\boldsymbol{O_k}}}{b_j^2}\right)};k=0,1;j=0,1,{\cdots},2m
\label{eq20}
\end{equation}

\begin{equation}
\boldsymbol{O_k} = 
\begin{cases}
\text{diag}\begin{bmatrix}
    o_{0} & 1-o_{0} & 0 & 0
\end{bmatrix},0<o_{0}<1; k=0 \\
\text{diag}\begin{bmatrix}
    0 & 0 & o_{1} & 1- o_{1}
\end{bmatrix},0<o_{1}<1; k=1
\label{eq21}
\end{cases}
\end{equation}
where $\boldsymbol{W}_* \in \mathbb{R}^{(2m+1)\times2}$ is the ideal network weights, and $\boldsymbol{W}_* = \begin{bmatrix} \boldsymbol{W_1}, & \boldsymbol{W_2} \end{bmatrix}$. $\boldsymbol{h}\left(\boldsymbol\cdot\right) \in \mathbb{R}^{2\times(2m+1)}$ is the Gaussian radial basis functions, and $\boldsymbol{h}\left(\boldsymbol\cdot\right) = \begin{bmatrix} \boldsymbol{h_1}; & \boldsymbol{h_2} \end{bmatrix}$. $\boldsymbol{\Gamma} = diag \begin{bmatrix} \Gamma_1, & \Gamma_2\end{bmatrix}$ is the weight scale coefficients of the two sub-networks. Calculation details of $\text{d2m}(\boldsymbol\cdot)$ can be seen in \eqref{eq19}. $\boldsymbol{c_j}$ and ${b_j}$ are the center vector and the width of the hidden layer, respectively. And
\begin{equation}
\boldsymbol{c_j} = \begin{bmatrix}\frac{j-m}{m} & \frac{j-m}{m} & \frac{j-m}{m} & \frac{j-m}{m}\end{bmatrix}^T
\notag
\end{equation}
$\boldsymbol{\chi}\in\mathbb{R}^{4\times1}$ represents the network's input matrix, while $o_0$ and $o_1$ are employed to regulate the significant weights of the input. Details about the input $\boldsymbol{\chi}$ will be provided below.

\begin{figure}[tb]
\centering
\includegraphics[width=7cm]{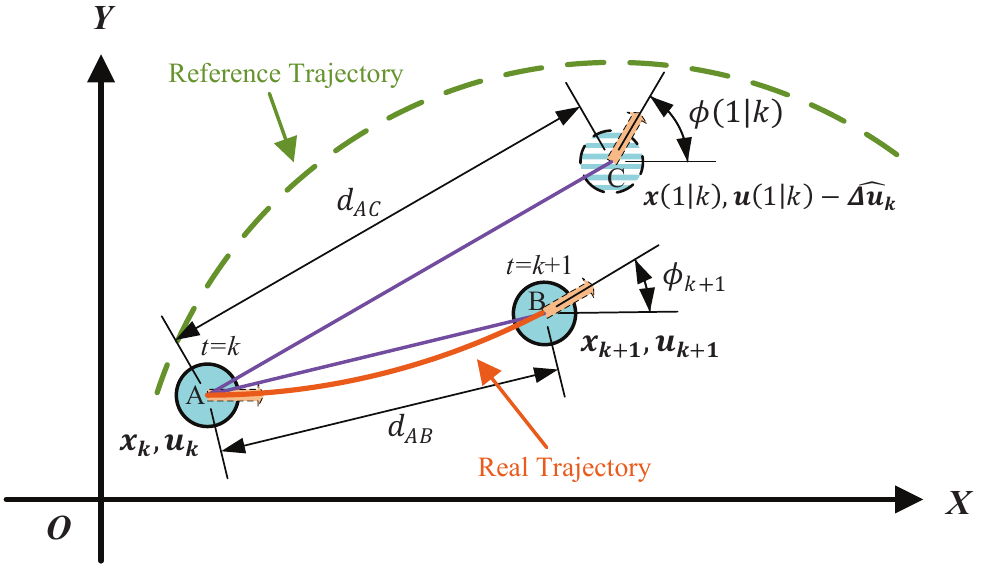}
\caption{Uncompensated uncertainties' impact in one control cycle. A is the robot's initial position at time k. An estimation is made at this moment that the estimated position in one single time is C, based on the given control instruction. However, the robot arrives B in time k+1 due to the uncompensated uncertainties.}
\label{fig8}
\end{figure}

The selection of input matrix $\boldsymbol{\chi}$ is essential for the RBFNN's fitting effect. To select the suitable input matrix, we may simulate the entire control cycle and identify key components. Assume the initial states and inputs of the system at time $k$ are $\boldsymbol{x_k}$ and $\boldsymbol{u_k}$, respectively. Additionally, the current estimated network weight and uncertainties at this moment are, respectively, $\boldsymbol{\hat{W}_k}$ and $\boldsymbol{\widehat{\Delta u}_k}=\begin{bmatrix}
    \widehat{\Delta v}_k & \widehat{\Delta q_r}_k
\end{bmatrix}$. And define the adaptive instruction planner's policy is $\boldsymbol{\Pi_{\hat{W}_k}}\left(\boldsymbol{x_k}\right)$. With initial states and inputs, we can predict the future states in the following N steps using $\boldsymbol{\Pi_{\hat{W}_k}}\left(\boldsymbol{x_k}\right)$, and the estimated states in the i steps after the moment k are defined as $\boldsymbol{{x}}\left(i|k\right), 1\leq i\leq N$. However, due to the discrepancy between $\boldsymbol{\hat{W}_k}$ and $\boldsymbol{{W}_k}$, the actual states in the next step $\boldsymbol{x_{k+1}}$ may not match the estimated value $\boldsymbol{{x}}\left(1|k\right)$. Fig.~\ref{fig8} demonstrates the above situations. Then let $\boldsymbol{E_e}$ represent the generalized uncertain state errors at time $k$ with respect to the estimated states $\boldsymbol{{x}}\left(1|k-1\right)$. 

If the step time ($\text{d}t$) is small enough, the path can be reduced to a straight line connecting two neighboring coordinate points. According to Fig.~\ref{fig8}, the difference between real and estimated moving distance $d_{AB}$ and $d_{AC}$ is connected to $ \Delta v-\widehat{\Delta v}_k $ while the yaw angle distance $\phi_{k+1}-{\phi}(1|k)$ is related to $\Delta q_r-\widehat{\Delta q_r}_k$, according to \eqref{eq2} and \eqref{eq17}. In addition, $\boldsymbol{u}(1|k)-\boldsymbol{\widehat{\Delta u}_k}$ should gradually converge to $\boldsymbol{u_{k+1}}$ based on \eqref{eq17}. Thus the appropriate input matrix $\boldsymbol{\chi}$ can be developed as follows.

\begin{equation}
\boldsymbol{\chi} = \text{MaxAbs}\left(\begin{bmatrix}
\partial \hat{d}_k & \partial \hat{v}_k & \partial \hat{\phi}_k & \partial \hat{q}_{rk}
\end{bmatrix}^T\right)
\label{eq23}
\end{equation}
where $\text{MaxAbs}(\cdot)$ is the max-abs normalization function and
\begin{equation}
\begin{aligned}
\partial \hat{d}_k =& \;\;d_{AB}-d_{AC} \notag\\
\partial \hat{v}_k =& \;\;v(0|k+1)-{{v}}(1|k)+\widehat{\Delta v}_k\notag\\
\partial \hat{\phi}_k =& \;\;\phi_{k+1}-{\phi}(1|k)\notag\\
\partial \hat{q}_{rk} =& \;\;q_r(0|k+1)-{{q}_r}(1|k)+\widehat{\Delta q}_{rk}\notag
\end{aligned}
\end{equation}

\subsection{Adaptive Instruction Planner: AN-MPC}

\begin{figure}[b]
\centering
\vspace{-0.3cm}
\subfigure[Actor-critic algorithm]
{
    \label{fig9:subfig:a} 
    \includegraphics[width=3cm]{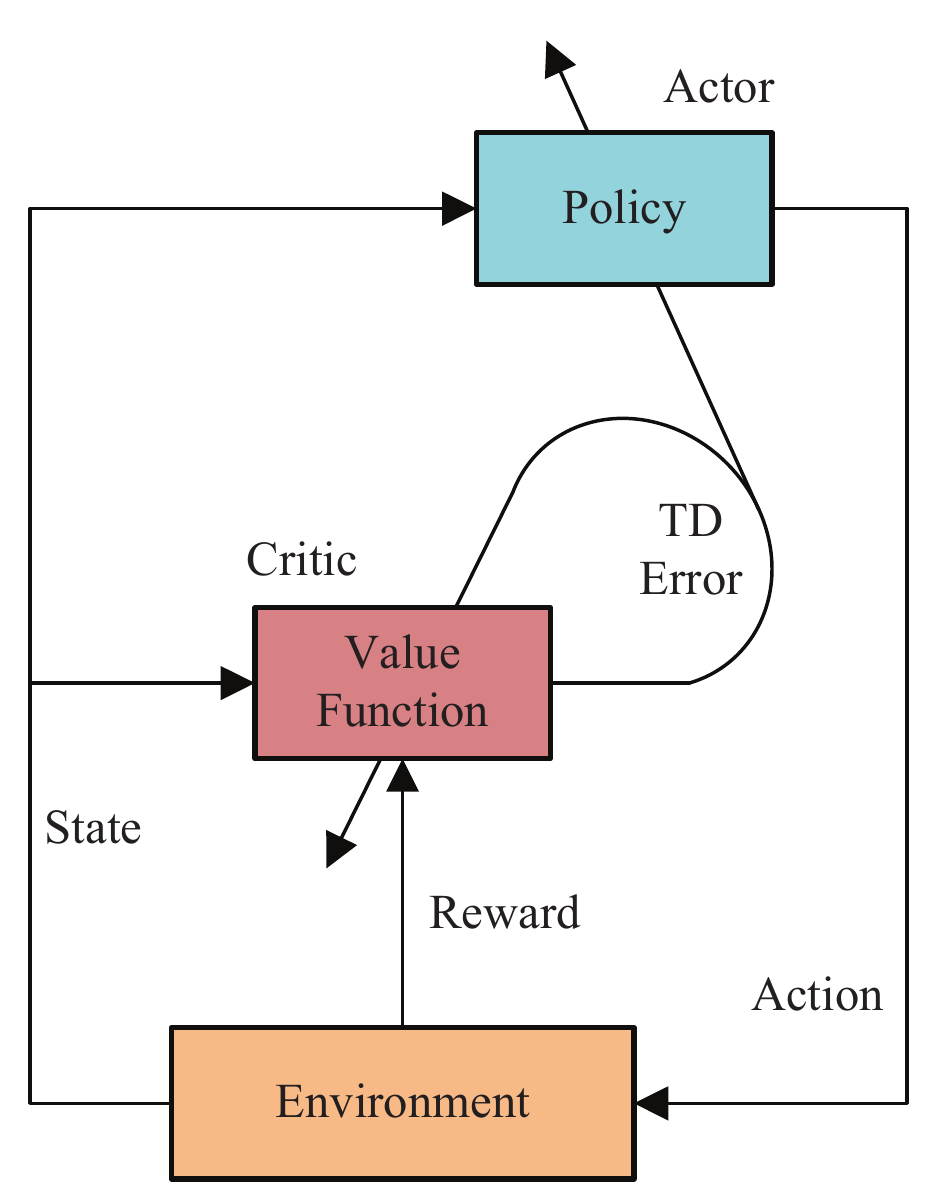}}
    \hspace{-0.1in}    
\subfigure[AN-MPC algorithm]
{
    \label{fig9:subfig:b} 
    \includegraphics[width=3cm]{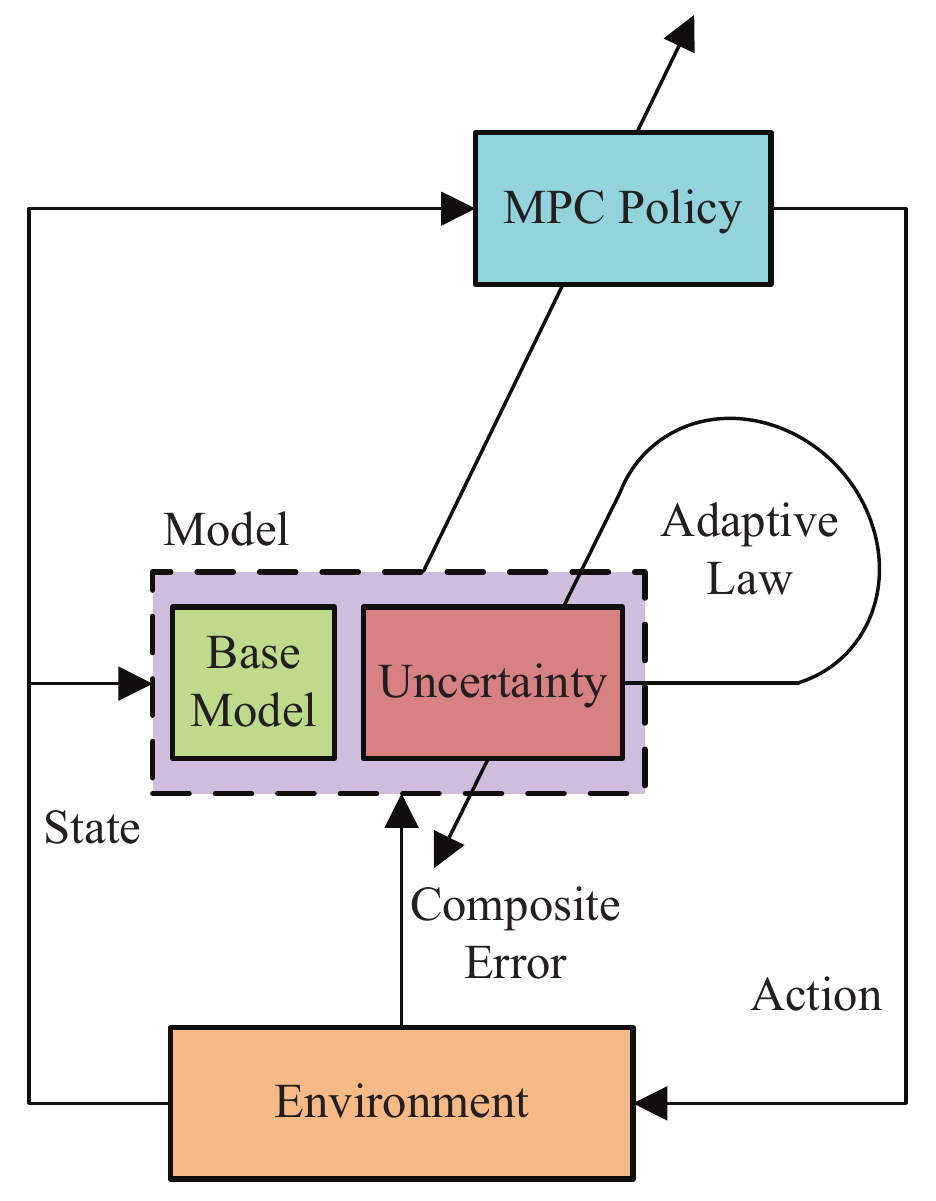}}
    \hspace{0in}
\caption{Comparison between the architecture of AN-MPC and actor-critic algorithm.}
\label{fig9}
\end{figure}

Prior to introducing the adaptive instruction planner, we must clarify two types of state errors. The first are the uncertain state errors $\boldsymbol{E_e}$ defined above, which will approach $\boldsymbol{0}$ when $\boldsymbol{\hat{W}_k}$ approaches its true value. And the second are the trajectory tracking state errors $\boldsymbol{E_r}$ with regard to the reference trajectory $\boldsymbol{x}_{ref}$. The uncertainties will be updated in each iteration, and then the planner will be applied to obtain the optimal command. Furthermore, we also define a composite error $\boldsymbol{E_c} = \gamma\boldsymbol{E_e} + \left(1-\gamma\right)\boldsymbol{E_r}, 0.5<\gamma<1.0$ to be applied in the adaptive portion. 

Additionally, $\boldsymbol{E_c}$ can be regarded as part of the reward function in RL, and the AN-MPC algorithm may have similarities with actor-critic algorithms \cite{rl10}. Fig.~\ref{fig9} displays a comparison between the architecture of AN-MPC and the actor-critic algorithm \cite{rl8}. The $\boldsymbol{E_e}$ portion and $\chi$ can assist in learning the true uncertainties, allowing the model used to update with each iteration and be as close as possible to the real model. Then the MPC portion in AN-MPC can calculate the optimal command based on the updated model, enhancing the quality of the solution to the trajectory tracking problem. The $\boldsymbol{E_r}$ part accounts for a small proportion but can accelerate the convergence of trajectory tracking.

Following the prior preparations, we propose the adaptive instruction planner AN-MPC which are detailed below.

\begin{align}
\min_{\boldsymbol{u}(\cdot)} & && J(\cdot)=\sum_{i=0}^N
\left\|\boldsymbol{x}-\boldsymbol{x}_{ref}\right\|^2_{\boldsymbol{Q}}+\sum_{i=0}^{N-1} \left\|\boldsymbol{u}-\boldsymbol{\widehat{\Delta u}_k}-\boldsymbol{u}_{ref}\right\|^2_{\boldsymbol{R}} 
\label{eq24}\\
\text{s.t.} & &&
\boldsymbol{\dot{x}}(i|k)=\boldsymbol{f}\left(\boldsymbol{x}(i|k),\;\; \boldsymbol{u}(i|k),\;\;\boldsymbol{\widehat{\Delta u}_k}\right),\tag{16a} \label{eq24a}\\
& && \boldsymbol{x}(0|k)=\boldsymbol{x_k},
\tag{16b} \label{eq24b}\\
& && \boldsymbol{x}(i|k)\in\left[
\boldsymbol{x}_{min},\;\; \boldsymbol{x}_{max}\right],\tag{16c} \label{eq24c}\\
& && \boldsymbol{u}(i|k)\in\left[
\boldsymbol{u}_{min}+\boldsymbol{\widehat{\Delta u}_k},\;\; \boldsymbol{u}_{max}+\boldsymbol{\widehat{\Delta u}_k}\right], \tag{16d} \label{eq24d} \\
& && G\left(\boldsymbol{x}(i|k),\;\; \boldsymbol{u}(i|k)\right)\leq 0, \tag{16e} \label{eq24e} \\
& && \boldsymbol{\widehat{\Delta u}_k} = \boldsymbol{\Gamma} \cdot \text{d2m}\left(\boldsymbol{h}\left(\boldsymbol{\chi}\right)\boldsymbol{\hat{W}}(k)\right),\tag{16f} \label{eq24f}\\
& && H_{clf}\left(\boldsymbol{x_k},\; \boldsymbol{u_k}, \; \boldsymbol{\hat{W}}(k),\; \boldsymbol{E_e},\;\boldsymbol{E_r},\;\boldsymbol{\dot{\hat{x}}}_{[-1]},\;\boldsymbol{\dot{x}}_{ref}
\right)\geq 0 \tag{16g} \label{eq24g}
\end{align}
where $N$ is the prediction horizon, ${\boldsymbol{Q}}$ and ${\boldsymbol{R}}$ are definite weighting coefficient matrix. $\boldsymbol{{x}}_{ref}=\begin{bmatrix} X_{ref} & Y_{ref} & \phi_{ref} \end{bmatrix}^T$ and $\boldsymbol{{u}}_{ref}=\begin{bmatrix} v_{ref} & q_{ref} \end{bmatrix}^T$ are reference trajectory and reference inputs, respectively. \eqref{eq24a} is the adaptive kinematic model which can also be seen in \eqref{eq17}. Inequality constraint \eqref{eq24e} is employed to indicate the boundary value problem of direct multi-shooting method. Initial states and inputs are defined as $\boldsymbol{x_k}$ and $\boldsymbol{u_k}$, respectively, which are obtained by sensors. And $\boldsymbol{\dot{\hat{x}}}_{[-1]} = \boldsymbol{\dot{x}}(1|k-1)$. $\boldsymbol{\dot{x}}_{ref}$ specifically refers to the derivative of the reference trajectory at time k. Moreover, the network weights $\boldsymbol{\hat{W}}(k)$ will be updated with each iteration according to the following adaptive law.

\begin{equation}
\dot{\hat{\boldsymbol{W}}}^T=\begin{bmatrix} \dot{\hat{\boldsymbol{W}}}_1^T \\ \dot{\hat{\boldsymbol{W}}}_2^T \end{bmatrix} 
= -\begin{bmatrix} \Gamma_1\boldsymbol{E_c}^T\boldsymbol{F_{u}^{'}} { \begin{bmatrix} \boldsymbol{h_1} & \boldsymbol O\end{bmatrix}}^T \\ \Gamma_2\boldsymbol{E_c}^T\boldsymbol{F_{u}^{'}} { \begin{bmatrix} \boldsymbol O & \boldsymbol{h_2}\end{bmatrix}}^T \end{bmatrix}
\label{eq25}
\end{equation}

\begin{equation}
\hat{\boldsymbol{W}}(k) = \hat{\boldsymbol{W}}(k-1) + \dot{\hat{\boldsymbol{W}}}^T\cdot \text{d}t
\label{eq26}
\end{equation}

The inequality constraint \eqref{eq24g} is the control Lyapunov function constraint, with the function $H_{clf}(\boldsymbol\cdot)$ as follows. 
\begin{equation}
\begin{aligned}
H_{clf} &= -\boldsymbol{E_c}^T\left[
        \boldsymbol{f}(\boldsymbol\cdot)-\boldsymbol{F_{u}^{'}}{\boldsymbol{\Gamma} \text{d2m} \left(\boldsymbol{h}\hat{\boldsymbol{W}}\right)}\right]+\gamma\boldsymbol{E_e}^T\boldsymbol{\dot{\hat{x}}}_{[-1]}\\
        &\;\;\;\;+(1-\gamma)\boldsymbol{E_r}^T\boldsymbol{\dot{x}}_{ref}-\frac{1}{2}\boldsymbol{E_c}^T\boldsymbol{K}\boldsymbol{E_c}
 \label{eq27}
 \end{aligned}
\end{equation}
where $\boldsymbol{K}\in\mathbb{R}^{3\times3}$ is the positive diagonal matrix.

The overall process of the AN-MPC algorithm is presented in \textbf{Algorithm~\ref{algorithm1}}.  
And as for now, we have obtained the new trajectory tracking framework ANMHH with the AN-MPC, HSMC and HTSMC. Moreover, experiments revealed that the convergence time of the adaptive process varies among terrains due to the varying magnitudes of uncertainties. We anticipate that when the amount of uncompensated uncertainties' absolute value $| \boldsymbol{\widehat{\Delta u}_k}-\boldsymbol{\Delta u}|$ is large, the adaptive speed will be sped up, and when it is small, the adaptive speed will be slowed down, allowing the algorithm to maintain a faster convergence speed and a higher level of stability. Therefore, we propose the variable step-size algorithm in the following subsection.

\IncMargin{1em}
\begin{algorithm} [tb]
\SetKwInOut{Input}{Input}\SetKwInOut{Output}{Output}
\SetKwInOut{Init}{Initialize}
\SetKwFunction{Env}{Env}
\SetKwFunction{CalculateRBFNNState}{GetRBFNNState}
\SetKwFunction{CalculateJacobi}{CalculateJacobi}
\SetKwFunction{AdaptiveMPC}{AdaptiveMPC}
\SetKwFunction{ReferDot}{ReferDerivate}
\SetKwFunction{EstimateDerivate}{EstimateDerivate}
\SetKwFunction{UpdateWeight}{UpdateWeight}
\SetKwFunction{push}{push}
\SetKwFunction{pop}{pop}
\SetKwFunction{size}{size}
\SetKwFunction{Filter}{Filter}
\caption{AN-MPC}
\label{algorithm1} 
\Input{Current real states and inputs from sensors: $\boldsymbol{x_{i}}$, $\boldsymbol{u_{i}}$\\
Reference trajectory: $\boldsymbol{x_r}$, $\boldsymbol{u_r}$}
\Output{Optimal command: $\boldsymbol{u_{o}}$}
\Init{Predictions of states and inputs: $\boldsymbol{x}$, $\boldsymbol{u}$\\ $\boldsymbol{E_c}$, $\boldsymbol{E_e}$, $\boldsymbol{E_r}$, $\boldsymbol{\hat{W}}$}
\newcommand\mycommfont[1]{\itshape\rmfamily\textcolor{RoyalBlue}{#1}}
\SetCommentSty{mycommfont}
\emph{Set hyperparameter values and initialize the control problem}\; 
\For{$k\leftarrow 0$ \KwTo $t$}{
    $\boldsymbol{\chi} \leftarrow \CalculateRBFNNState(\boldsymbol{x_i}, \boldsymbol{u_i}, \boldsymbol{x}[1], \boldsymbol{u}[0])$\tcp*[r]{Eq.(15) and subsection \uppercase\expandafter{\romannumeral3}.A}
        $\boldsymbol{\hat{f}_{u}^{'}} \leftarrow \CalculateJacobi(\boldsymbol{x_i},\boldsymbol{u_i})$\;
    $\boldsymbol{\hat{W}}\leftarrow \UpdateWeight(\boldsymbol{\hat{W}},\boldsymbol{h}(\boldsymbol\chi),\boldsymbol{E_c},\boldsymbol{\hat{f}_{u}^{'}})$\tcp*[r]{Update RBFNN's weights using Eq.(17) and Eq(18) before estimating the uncertainty}    
    $\boldsymbol{\widehat{\Delta u}} = \boldsymbol{\Gamma} \cdot \text{d2m}\left(\boldsymbol{h}\left(\boldsymbol{\chi}\right)\boldsymbol{\hat{W}}\right)$\tcp*[r]{Estimate uncertainty}
    $\boldsymbol{\dot{x}_r^{k}} \leftarrow \ReferDot(\boldsymbol{x_r}[k],\boldsymbol{u_r}[k])$;
    $\boldsymbol{\dot{\hat{x}}}_{[-1]} \leftarrow \EstimateDerivate(\boldsymbol{x}[1], \boldsymbol{u}[1])$\tcp*[r]{Eq.(10)}
    $\boldsymbol{\Psi} \leftarrow (\boldsymbol{x_i}, \boldsymbol{u_i}, \boldsymbol{x_r}, \boldsymbol{u_r}, \boldsymbol{\widehat{\Delta u}}, \boldsymbol{\hat{W}},$\\$ \qquad \qquad  \boldsymbol{E_e}, \boldsymbol{E_r}, \boldsymbol{\dot{\hat{x}}}_{[-1]},\boldsymbol{\dot{x}_r^{k}},\boldsymbol{\hat{f}_{u}^{'}})$\;
    $(\boldsymbol{x},\boldsymbol{u}) \leftarrow$ Solve $\AdaptiveMPC(\boldsymbol{\Psi})$ \tcp*[r]{Eq.(16)}
    $\boldsymbol{u_o}=\boldsymbol{u}[0]$\tcp*[r]{Optimal command}
    $\boldsymbol{x_{i}},\boldsymbol{u_{i}} \leftarrow \Env(\boldsymbol{u_o})$ \tcp*[r]{Apply and observe}
    $\boldsymbol{E_e} = \boldsymbol{x_i}-\boldsymbol{x}[1] , \boldsymbol{E_r} = \boldsymbol{x_i}-\boldsymbol{x_r}[k+1]$\;
    $\boldsymbol{E_c}=\gamma\boldsymbol{E_e} + \left(1-\gamma\right)\boldsymbol{E_r}$;
 }
\end{algorithm}
\DecMargin{1em} 

\subsection{Variable Step-size Algorithm and VAN-MPC}
We designed the variable step-size algorithm for the reasons stated previously. And we discovered that the $\boldsymbol\Gamma$ has similar properties to the step-size in machine learning, i.e., when $\boldsymbol\Gamma$ is large, the learning speed of the uncertainties is fast but prone to oscillation and divergence, and when $\boldsymbol\Gamma$ is small, the learning speed is slow but relatively stable. On the basis of this information, we plan to construct a variable step-size with $\boldsymbol\Gamma$ and modify it to account for uncompensated uncertainty. The problem then becomes determining which observations can characterize the size of the uncompensated uncertainty.

The answer may be found in RBFNN since the states in RBFNN's input matrix $\boldsymbol{\chi}$ is closely connected with the uncompensated uncertainties. In addition, RBFNN depends on the distance between $\boldsymbol{\chi}$ and the center vector $\boldsymbol{c_j}$. We then divide the size of the uncompensated uncertainties into discrete $2m+1$ levels, and the number of the center vector with the shortest distance represents the size of the uncompensated uncertainties. Define $\boldsymbol{\zeta_k} = \begin{bmatrix}
    \zeta_{k}^{v} & \zeta_{k}^{q_r} 
\end{bmatrix}$ to represent the level of uncompensated uncertainties $\boldsymbol{\widehat{\Delta u}_k}-\boldsymbol{\Delta u}$ at time $k$, and:
\begin{equation}
\begin{cases}
\zeta_{k}^{v} &= \left[\arg\min\limits_{j\in 0,1,\cdots,2m} \left(\Vert\boldsymbol{\chi}-\boldsymbol{c_{j}}\Vert^2_{\boldsymbol{O_0}}\right)-m\right]/m\\
\zeta_{k}^{q_r} &= \left[\arg\min\limits_{j\in 0,1,\cdots,2m} \left(\Vert\boldsymbol{\chi}-\boldsymbol{c_{j}}\Vert^2_{\boldsymbol{O_1}}\right)-m\right]/m
\label{eq28}
\end{cases}
\end{equation}

However, this parameter $\boldsymbol{\zeta_k}$ cannot be utilized directly in the variable step-size algorithm since the introduction of new gradient propagation may impact the adaption laws and increase the adaptive law's index. To breakdown the gradient propagation, we build a replay buffer $\mathcal{B}$ to record the newest multiple $\boldsymbol{\zeta_k}$ and use $\overline{\boldsymbol{\zeta_k}}$ instead.
\begin{equation}
\overline{\boldsymbol{\zeta_k}}=\begin{bmatrix}
\overline{\zeta_k^v} & \overline{\zeta_k^{q_r}}
\end{bmatrix} = \mathbb{E}\left[ \boldsymbol{\zeta_i}| \boldsymbol{\zeta_i}\in \mathcal{B}\right]
\label{eq29}
\end{equation}

\begin{figure}[tb]
\centering
\includegraphics[width=5cm]{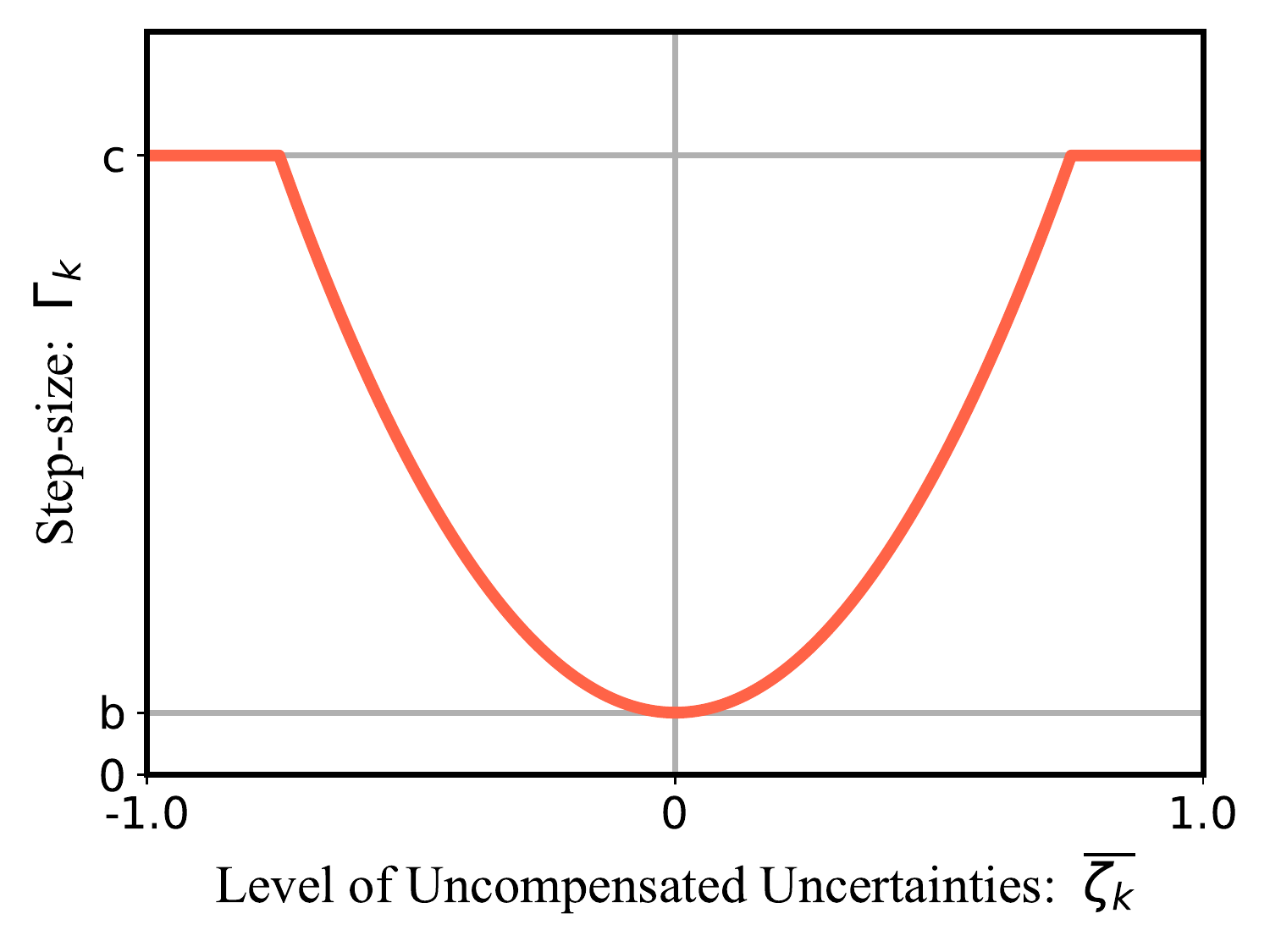}
\caption{The variable step-size algorithm, where $\Gamma_k$ represents $\Gamma_1$ and $\Gamma_2$. $\overline{{\zeta_k}}$ represents $\overline{{\zeta_k^v}}$ and $\overline{{\zeta_k^{q_r}}}$. And $b$ stands for $b_v$ and $b_{q_r}$, while $c$ stands for $c_v$ and $c_{q_r}$.}
\label{fig10}
\vspace{-0.2cm}
\end{figure}

As we mentioned above, we anticipate that the amount of uncompensated uncertainties' absolute value $|\boldsymbol{\widehat{\Delta u}_k}-\boldsymbol{\Delta u}|$ has a positive correlation with the step-size. Since $\overline{\boldsymbol{\zeta_k}}$ is employed to represent the amount of uncompensated uncertainties, the step-size $\boldsymbol{\Gamma}$ should be proportional to the $\overline{\boldsymbol{\zeta_k}}$'s absolute value $|\overline{\boldsymbol{\zeta_k}}|$. Actually, there are numerous functions that meet this requirement. We experimented with several functions before settling on quadratic functions for the variable step-size algorithm. And the algorithm we employ is presented below.
\begin{equation}
\begin{cases}
\boldsymbol{\Gamma} = diag \begin{bmatrix} \Gamma_1, & \Gamma_2\end{bmatrix}\\
 \Gamma_1 = \min\left(a_v\cdot\overline{{\zeta_k^v}}^2+b_v,\;\; c_v\right)\\
 \Gamma_2 = \min\left(a_{q_r}\cdot\overline{{\zeta_k^{q_r}}}^2+b_{q_r},\;\; c_{q_r}\right)
\label{eq30}
\end{cases}
\end{equation}
where $a_v$, $b_v$, $c_v$ and $a_{q_r}$, $b_{q_r}$, $c_{q_r}$ are both positive values that can influence the variation range and speed of the step-size. 
The schematic diagram of the variable step-size algorithm is shown in Fig.~\ref{fig10}.

Replacing the constant value $\boldsymbol{\Gamma}$ in AN-MPC with the variable step-size algorithm in \eqref{eq30}, we finally obtain the more efficient and robust instruction planner VAN-MPC. \textbf{Algorithm~\ref{algorithm2}} describes the overall process of the VAN-MPC algorithm. Cooperating with the previous velocity controller HSMC and direction controller HTSMC, the multi-terrain trajectory tracking control framework VANMHH is accomplished. Compared with MHH, multi-terrain control purpose is achieved just by optimizing a single component (the instruction planner), which is considerably very convenient. 

\begin{figure*}[tb]
\centering
\subfigure[Flat Tiled Floor]
{
    \label{fig11:subfig:a} 
    \includegraphics[width=3cm]{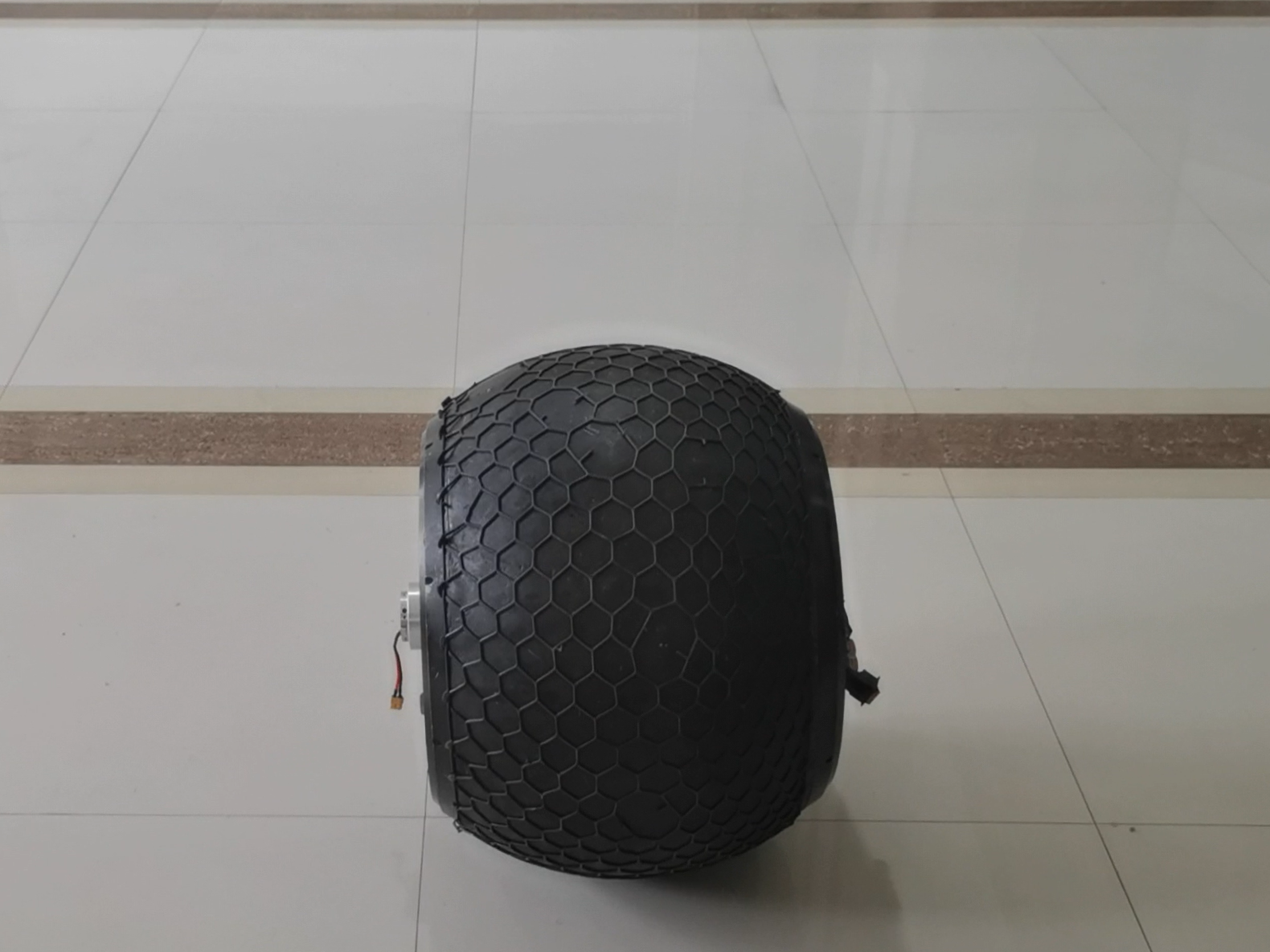}}
    \hspace{0.1in}    
\subfigure[Rubber Ground]
{
    \label{fig11:subfig:b} 
    \includegraphics[width=3cm]{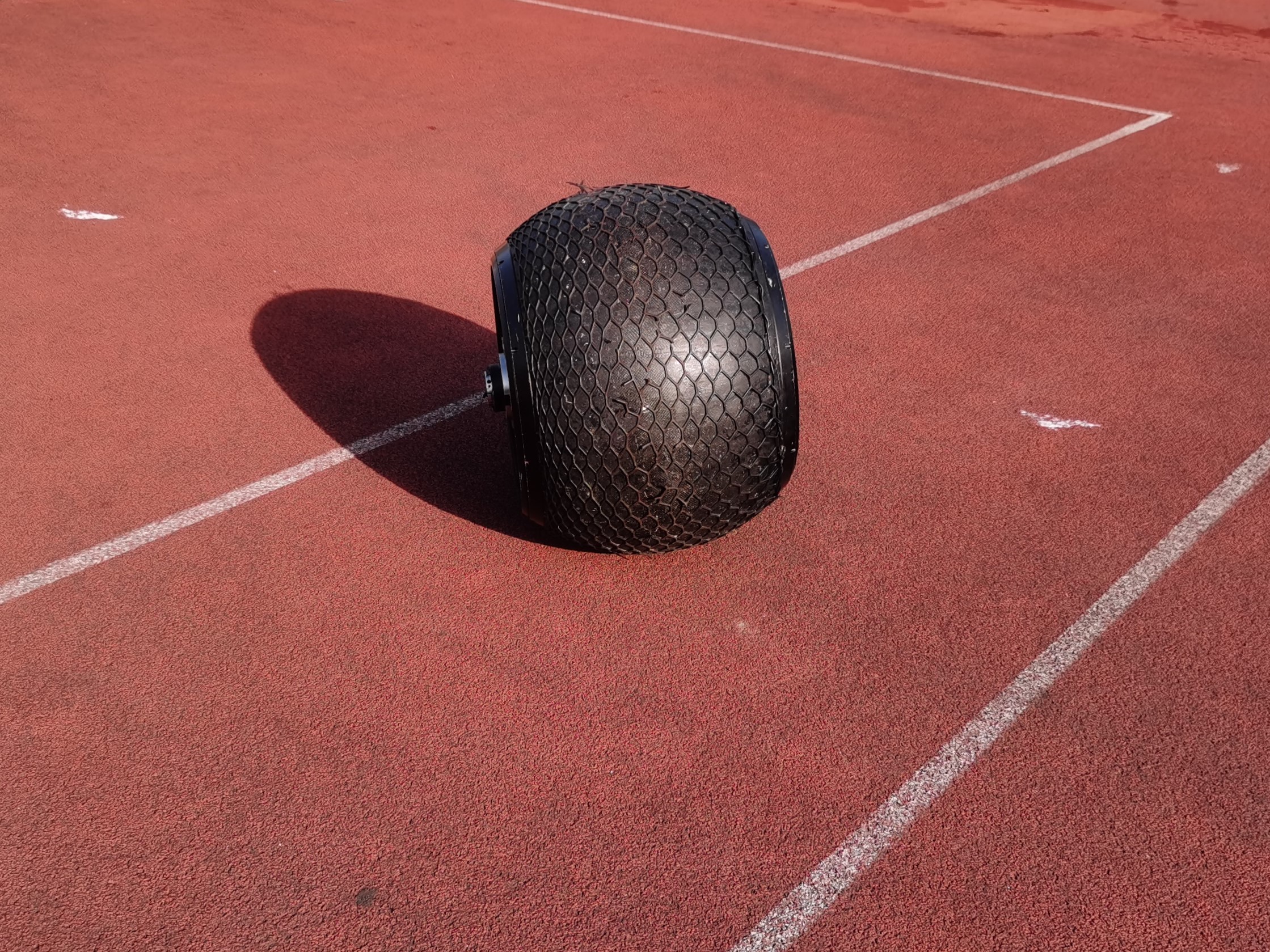}}
    \hspace{0.1in}
\subfigure[Ground with Hollow Tiles]
{
    \label{fig11:subfig:c}
    \includegraphics[width=3cm]{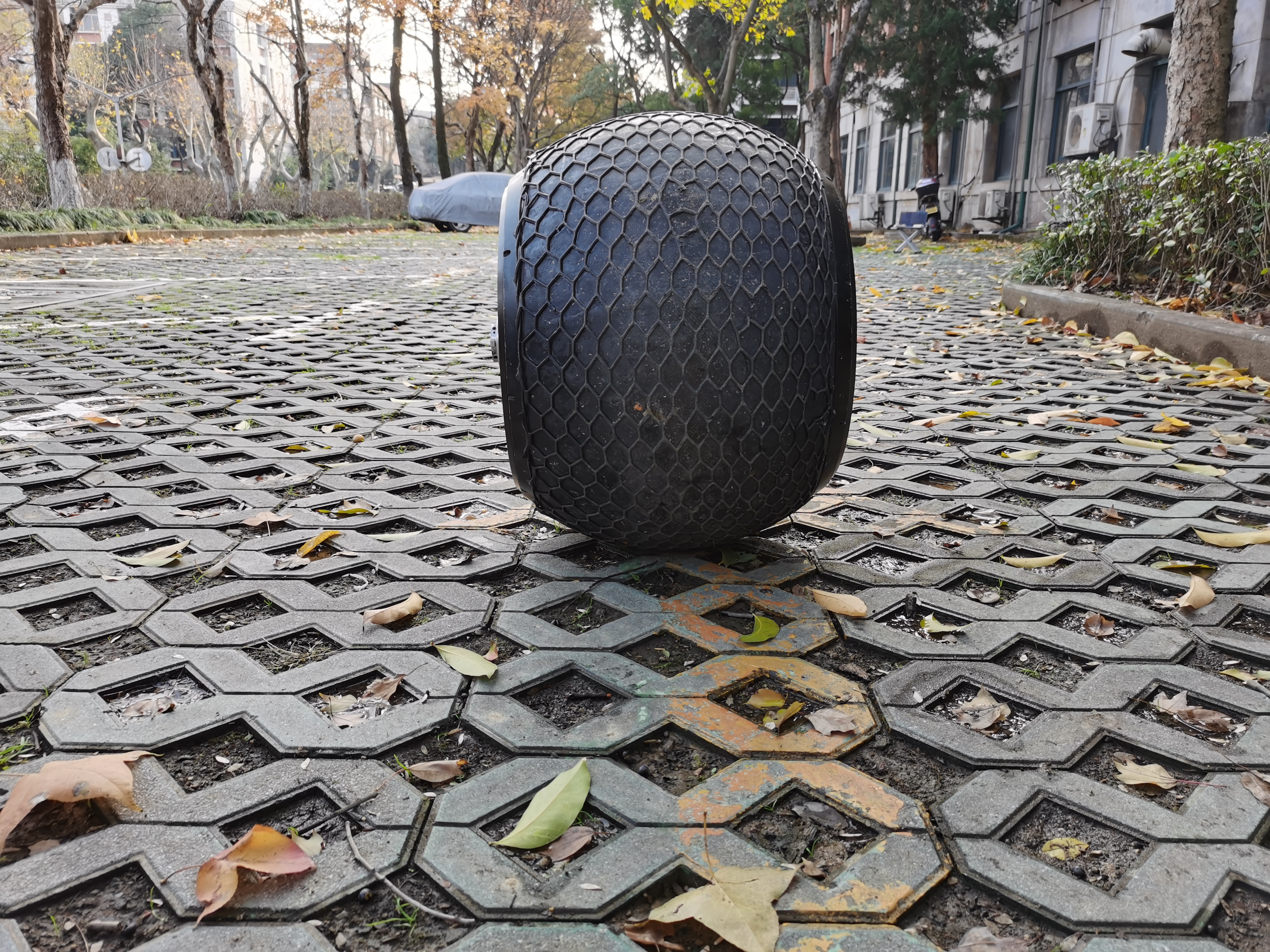}}
    \hspace{0.1in}
\subfigure[Grass]
{
    \label{fig11:subfig:d} 
    \includegraphics[width=3cm]{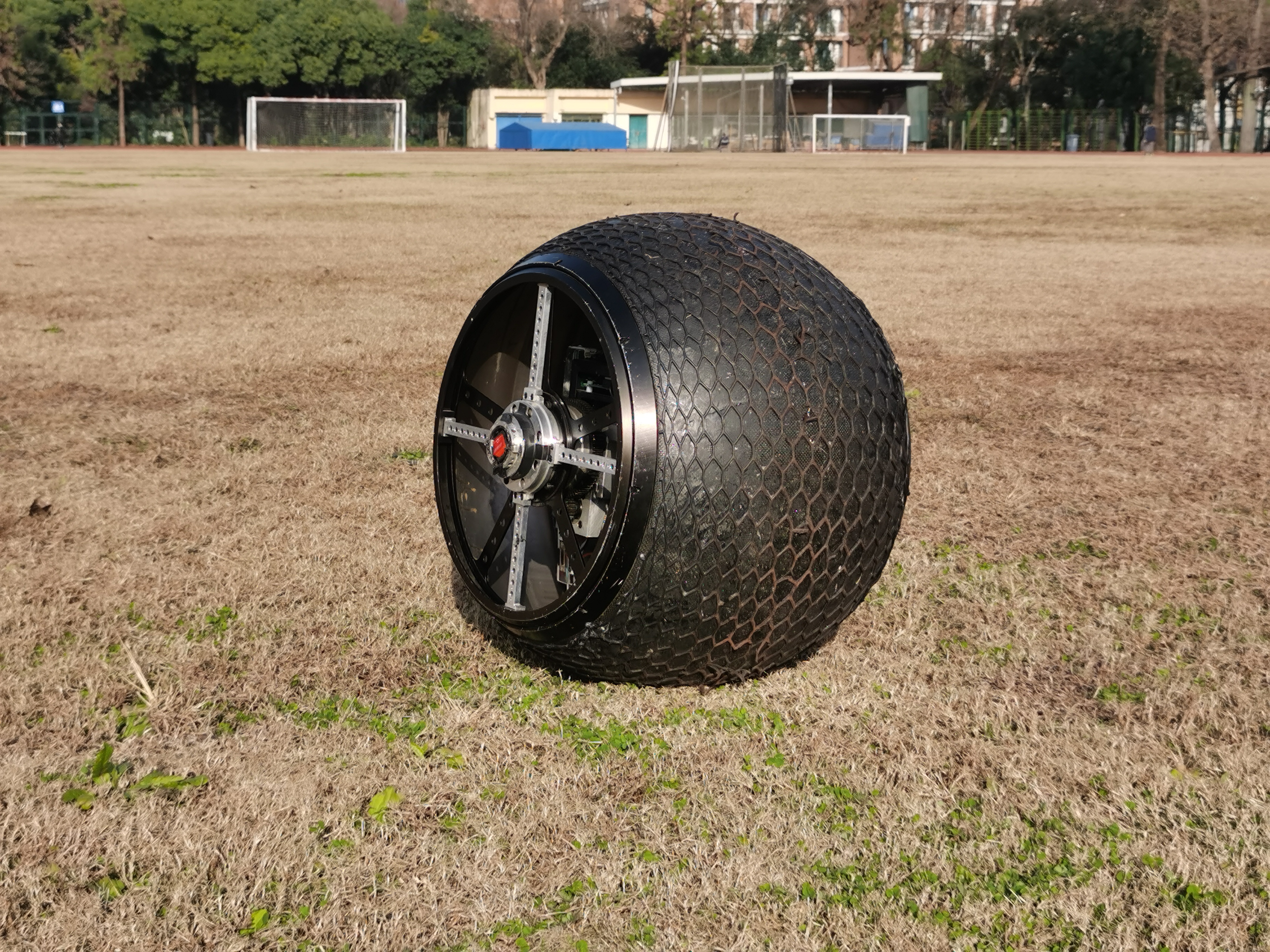}}
    \hspace{0.1in}
\subfigure[Varied Terrain]
{
    \label{fig11:subfig:e} 
    \includegraphics[width=3cm]{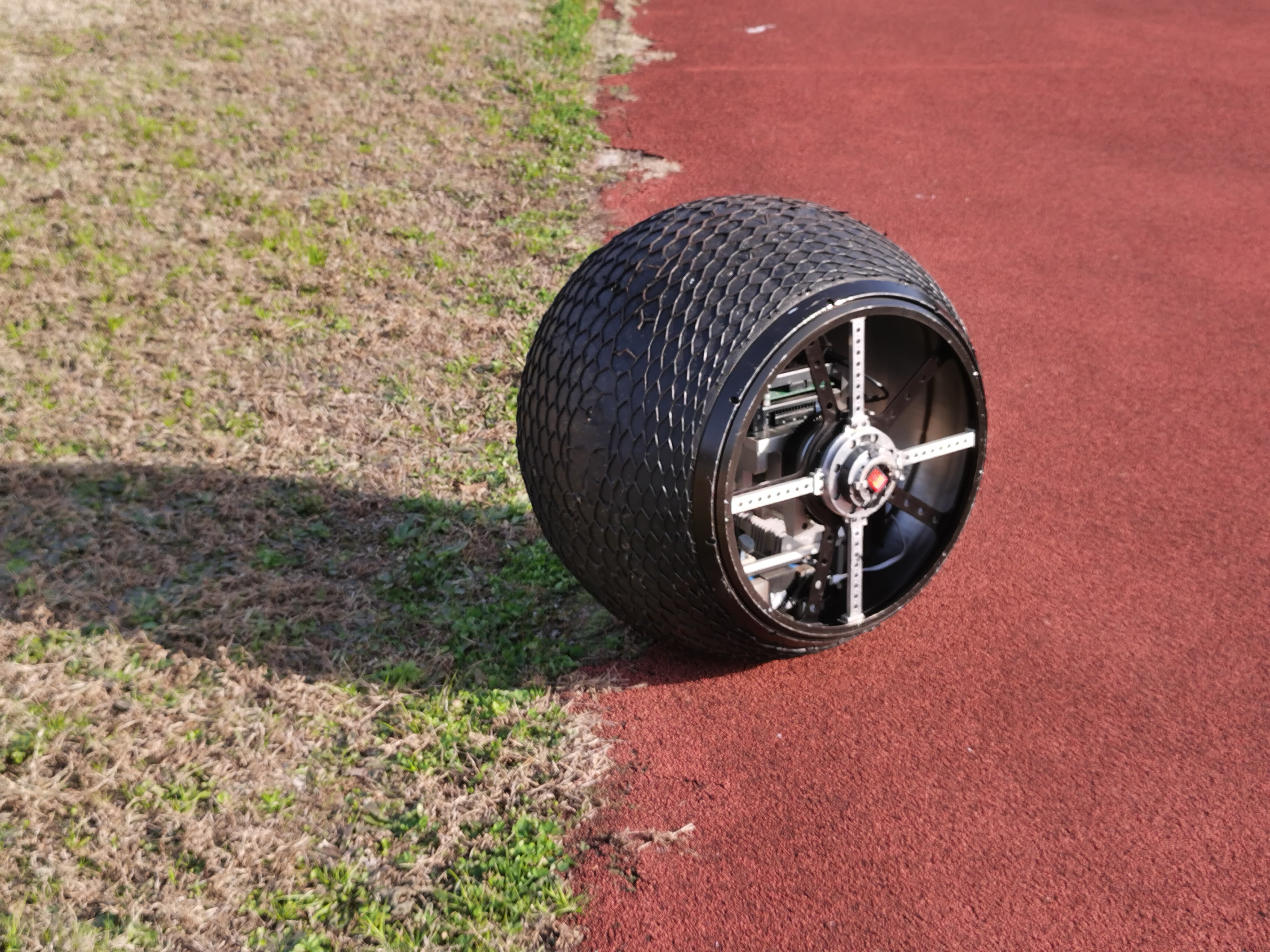}}
\caption{Terrains for experiments.}
\vspace{-0.5cm}
\label{fig11}
\end{figure*}

\subsection{Stability Analysis}
The ideal network weights are $\boldsymbol{W}_*$ which can perfectly match the uncertainties $\boldsymbol{\Delta u}$, whereas the estimated network weights are $\boldsymbol{\hat{W}}$. Now we define weights' errors $\widetilde{\boldsymbol{W}}=\boldsymbol{\hat{W}}-\boldsymbol{W}_*$. Construct the Lyapunov-like function $V\left(\boldsymbol{E_e}, \boldsymbol{E_r}, \widetilde{\boldsymbol{W}}\right)$ and a quadratic equation $Q\left(\boldsymbol{E_c}\right)$ which can be seen as follows:
\begin{equation}
V(\boldsymbol\cdot)=\frac{\gamma}{2}\boldsymbol{E_e}^T\boldsymbol{E_e}+\frac{1-\gamma}{2}\boldsymbol{E_r}^T\boldsymbol{E_r}+\frac{1}{2}\text{tr}\left(\widetilde{\boldsymbol{W}}^T\widetilde{\boldsymbol{W}}\right)
\label{eq31}
\end{equation}

\begin{equation}
Q(\cdot)= \frac{1}{2}\boldsymbol{E_c}^T\boldsymbol{K}\boldsymbol{E_c}
\label{eq32}
\end{equation}

Obviously $Q\left(\boldsymbol{0}\right)$ = 0 and $\lim_{\|\boldsymbol{E_r}\|\rightarrow\infty}Q\left(\boldsymbol{E_r}\right)=\infty$. In addition, $\boldsymbol{E_c} = \gamma\boldsymbol{E_e}+\left(1-\gamma\right)\boldsymbol{E_r}$, and $V\left(\boldsymbol\cdot\right)$ is the Lyapunov-like function. According to \cite{adap3, bib38}, if $\dot{V}\leq -Q$, $\boldsymbol{E_e}$ and $\boldsymbol{E_r}$ will gradually converges to $\boldsymbol{0}$. $\boldsymbol{E_e}$ converges to $\boldsymbol{0}$ also means that the uncertainties have been fully compensated according to \uppercase\expandafter{\romannumeral3}.A and \uppercase\expandafter{\romannumeral3}.B. Moreover, the inclusion of $\boldsymbol{E_r}$ in $V$ and $Q$ also aids in accelerating the convergence of itself. The inequality equation $\dot{V}\leq -Q$ will be demonstrated below. 

For the sake of brevity, define $\boldsymbol{I_{1,0}}^T=\begin{bmatrix} 1 & 0\end{bmatrix}$ and $\boldsymbol{I_{0,1}}^T=\begin{bmatrix} 0 & 1\end{bmatrix}$. Then we can get the derivative of $V$ as follows:
\begin{equation}
\begin{aligned}
\dot{V} &= \gamma\boldsymbol{E_e}^T{\boldsymbol{\dot{E}_e}}+(1-\gamma)\boldsymbol{E_r}^T{\boldsymbol{\dot{E}_r}}+\text{tr}\left(\widetilde{\boldsymbol{W}}^T\dot{\widetilde{\boldsymbol{W}}}\right)\\
        &= \boldsymbol{E_c}^T\left[
        f(\cdot)-\boldsymbol{F_{u}^{'}} {\boldsymbol{\Gamma}\text{d2m} \left(\boldsymbol{h}\boldsymbol{W}_*\right)}\right]-(1-\gamma)\boldsymbol{E_r}^T\boldsymbol{\dot{x}}_{ref}\\
        &\;\;\;\;-\gamma\boldsymbol{E_e}^T\boldsymbol{\dot{\hat{x}}}_{[-1]}+ \left( \boldsymbol{I_{1,0}}^T \widetilde{\boldsymbol{W}}^T\dot{\hat{\boldsymbol{W}}} \boldsymbol{I_{1,0}} + \boldsymbol{I_{0,1}}^T \widetilde{\boldsymbol{W}}^T\dot{\hat{\boldsymbol{W}}} \boldsymbol{I_{0,1}}\right)    \\
        &= \boldsymbol{E_c}^T\left[
        f(\cdot)-\boldsymbol{F_{u}^{'}}{\boldsymbol{\Gamma}\text{d2m}\left(\boldsymbol{h}\hat{\boldsymbol{W}}\right)}\right]-(1-\gamma)\boldsymbol{E_r}^T\boldsymbol{\dot{x}}_{ref}\\
        &\;\;\;\;-\gamma\boldsymbol{E_e}^T\boldsymbol{\dot{\hat{x}}}_{[-1]}+ \left( \boldsymbol{E_c}^T \boldsymbol{F_{u}^{'}}{\boldsymbol{\Gamma} \begin{bmatrix} 1 & 0 \\ 0 & 0\end{bmatrix} \boldsymbol h } + \boldsymbol{I_{1,0}}^T\dot{\hat{\boldsymbol{W}}}^T \right)\widetilde{\boldsymbol{W}}\boldsymbol{I_{1,0}}\\
        &\;\;\;\;+ \left( \boldsymbol{E_c}^T \boldsymbol{F_{u}^{'}}{\boldsymbol{\Gamma} \begin{bmatrix} 0 & 0 \\ 0 & 1\end{bmatrix} \boldsymbol h } + \boldsymbol{I_{0,1}}^T\dot{\hat{\boldsymbol{W}}}^T \right)\widetilde{\boldsymbol{W}}\boldsymbol{I_{0,1}}\\ 
        &= \boldsymbol{E_c}^T\left[
        f(\cdot)-\boldsymbol{F_{u}^{'}}{\boldsymbol{\Gamma} \text{d2m} \left(\boldsymbol{h}\hat{\boldsymbol{W}}\right)}\right]-(1-\gamma)\boldsymbol{E_r}^T\boldsymbol{\dot{x}}_{ref}\\
        &\;\;\;\;-\gamma\boldsymbol{E_e}^T\boldsymbol{\dot{\hat{x}}}_{[-1]}+ \left( \Gamma_1\boldsymbol{E_c}^T \boldsymbol{F_{u}^{'}} { \begin{bmatrix} \boldsymbol{h_1} \\ \boldsymbol O\end{bmatrix}} + \dot{\hat{\boldsymbol{W}}}_1^T \right)\widetilde{\boldsymbol{W}}_1\\
        &\;\;\;\;+ \left( \Gamma_2\boldsymbol{E_c}^T \boldsymbol{F_{u}^{'}} { \begin{bmatrix} \boldsymbol O \\ \boldsymbol{h_2}\end{bmatrix}} + \dot{\hat{\boldsymbol{W}}}_2^T \right)\widetilde{\boldsymbol{W}}_2
\end{aligned}
\label{eq33}
\end{equation}

Substitute the adaptive law \eqref{eq25} into \eqref{eq33}, and we can get:

\begin{equation}
\begin{aligned}
\dot{V} &=          \boldsymbol{E_c}^T\left[        \boldsymbol{f}(\boldsymbol\cdot)-\boldsymbol{F_{u}^{'}}{\boldsymbol{\Gamma} \text{d2m} \left(\boldsymbol{h}\hat{\boldsymbol{W}}\right)}\right]-(1-\gamma)\boldsymbol{E_r}^T\boldsymbol{\dot{x}}_{ref}\\
&\;\;\;\;-\gamma\boldsymbol{E_e}^T\boldsymbol{\dot{\hat{x}}}_{[-1]}
\end{aligned}
\label{eq34}
\end{equation}

At this moment, the equation $\dot{V}\leq-Q$ is equivalent to the inequality constraint \eqref{eq24g} and \eqref{eq27}. After adding this inequality constraint to the \eqref{eq24}, stability is guaranteed during control process.

\IncMargin{1em}
\begin{algorithm} [tb]
\SetKwInOut{Input}{Input}\SetKwInOut{Output}{Output}
\SetKwInOut{Init}{Initialize}
\SetKwFunction{Env}{Env}
\SetKwFunction{CalculateRBFNNState}{GetRBFNNState}
\SetKwFunction{CalculateJacobi}{CalculateJacobi}
\SetKwFunction{AdaptiveMPC}{AdaptiveMPC}
\SetKwFunction{ReferDot}{ReferDerivate}
\SetKwFunction{EstimateDerivate}{EstimateDerivate}
\SetKwFunction{UpdateWeight}{UpdateWeight}
\SetKwFunction{UpdateStepSize}{UpdateStepSize}
\SetKwFunction{push}{push}
\SetKwFunction{pop}{pop}
\SetKwFunction{size}{size}
\SetKwFunction{UncertaintySize}{UncertaintySize}
\caption{VAN-MPC}
\label{algorithm2} 
\Input{Current real states and inputs from sensors: $\boldsymbol{x_{i}}$, $\boldsymbol{u_{i}}$\\
Reference trajectory: $\boldsymbol{x_r}$, $\boldsymbol{u_r}$}
\Output{Optimal command: $\boldsymbol{u_{o}}$}
\Init{Predictions of states and inputs: $\boldsymbol{x}$, $\boldsymbol{u}$\\ $\boldsymbol{E_c}$, $\boldsymbol{E_e}$, $\boldsymbol{E_r}$, $\boldsymbol{\hat{W}}$, $\boldsymbol{\Gamma}$,
 and queue buffer $\mathcal{B}$}
\newcommand\mycommfont[1]{\itshape\rmfamily\textcolor{RoyalBlue}{#1}}
\SetCommentSty{mycommfont}
\emph{Set hyperparameter values and initialize the control problem}\; 
\For{$k\leftarrow 0$ \KwTo $t$}{
    $\boldsymbol{\chi} \leftarrow \CalculateRBFNNState(\boldsymbol{x_i}, \boldsymbol{u_i}, \boldsymbol{x}[1], \boldsymbol{u}[0])$\tcp*[r]{Eq.(15) and subsection \uppercase\expandafter{\romannumeral3}.A}
    $\boldsymbol{\zeta_k} \leftarrow \UncertaintySize(\boldsymbol{\chi})$ \tcp*[r]{Eq.(20)}
    $\mathcal{B}. \push(\boldsymbol{\boldsymbol{\zeta_k}}$)\;
    \If{$\mathcal{B}.\size() \geq l$}{$\mathcal{B}. \pop()$\;}
    $\overline{\boldsymbol{\zeta_k}} = \mathbb{E}\left[ \boldsymbol{\zeta_i}| \boldsymbol{\zeta_i}\in \mathcal{B}\right]$\tcp*[r]{Eq.(21)}
    $\boldsymbol{\Gamma} \leftarrow \UpdateStepSize(\overline{\boldsymbol{\zeta_k}})$\tcp*[r]{Eq.(22)}
    Lines 4 to 15 of the algorithm AN-MPC\;
 }
\end{algorithm}
\DecMargin{1em} 

\section{EXPERIMENTAL RESULTS}

\begin{figure}[b]
\vspace{-0.3cm}
\centering
\includegraphics[width=6cm]{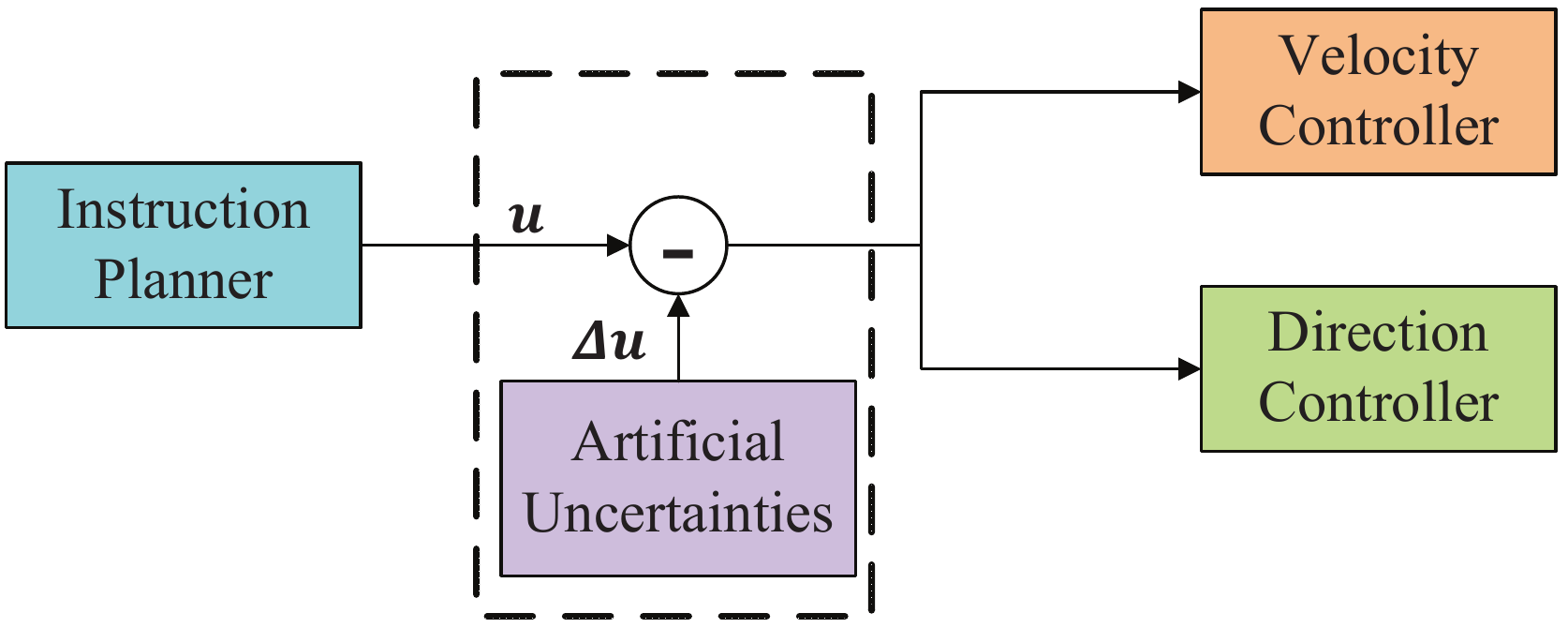}
\caption{Insert artificial uncertainties into trajectory tracking experiments to observe the uncertainties' convergence process. The area enclosed by the dashed line represents the module of artificial uncertainties inserted during the experiments in \uppercase\expandafter{\romannumeral4}.A. $\boldsymbol{\Delta u}=\begin{bmatrix}
    \Delta v & \Delta q_r
\end{bmatrix}$ represents the true value of the uncertainties.}
\label{fig12}
\end{figure}

In this section, we pretend to prove that our instruction planner VAN-MPC and trajectory tracking framework VANMHH can function efficiently on multiple terrains without prior knowledge of the terrains. Three more planners and their tracking frameworks are used for comparison, i.e., MPC and its framework MHH, two planners AN-MPC with large or small step-size, and their framework ANMHH. In addition, the spherical robot's base model is obtained on a flat tiled floor in Fig.~\ref{fig11:subfig:a}. All the algorithms run in concurrent threads on a mini PC (Intel i7-8559U, 2.70 GHz, Quad-core 64-bit). Moreover, the control frequency is 50 Hz for HSMC and HTSMC, and is 10 Hz for the planners. 

Five terrains are selected as shown in Fig.~\ref{fig11}. In subsection A, the robot will track a sine-wave trajectory on the flat tiled floor, where artificial uncertainties are introduced to quantify the effect of the algorithms. The following three terrains are used for tracking experiments in subsection B. And on the last terrain, the robot transits from rubber ground to grass while tracking a Lemniscate of Gerono. 
The variable step-size algorithm's hyperparameter values for VAN-MPC is $(a_v,b_v, c_v;a_{qr}, b_{qr}, c_{qr}) = (8, 0.3, 1.5; 0.3, 0.06, 0.15)$, while the constant step-size for AN-MPC (small step-size) and AN-MPC (large step-size) is $\boldsymbol{\Gamma}=(0.5, 0.1)$ and $\boldsymbol{\Gamma}=(1.0, 0.1)$.
Furthermore, CasADi framework and IPOPT are employed to solve the nonlinear program \cite{casadi, ipopt}.

\begin{figure}[t]
\setlength{\abovecaptionskip}{-2pt}
\centering
\subfigure[Real Path]
{
    \label{fig13:subfig:a} 
    \includegraphics[width=3cm]{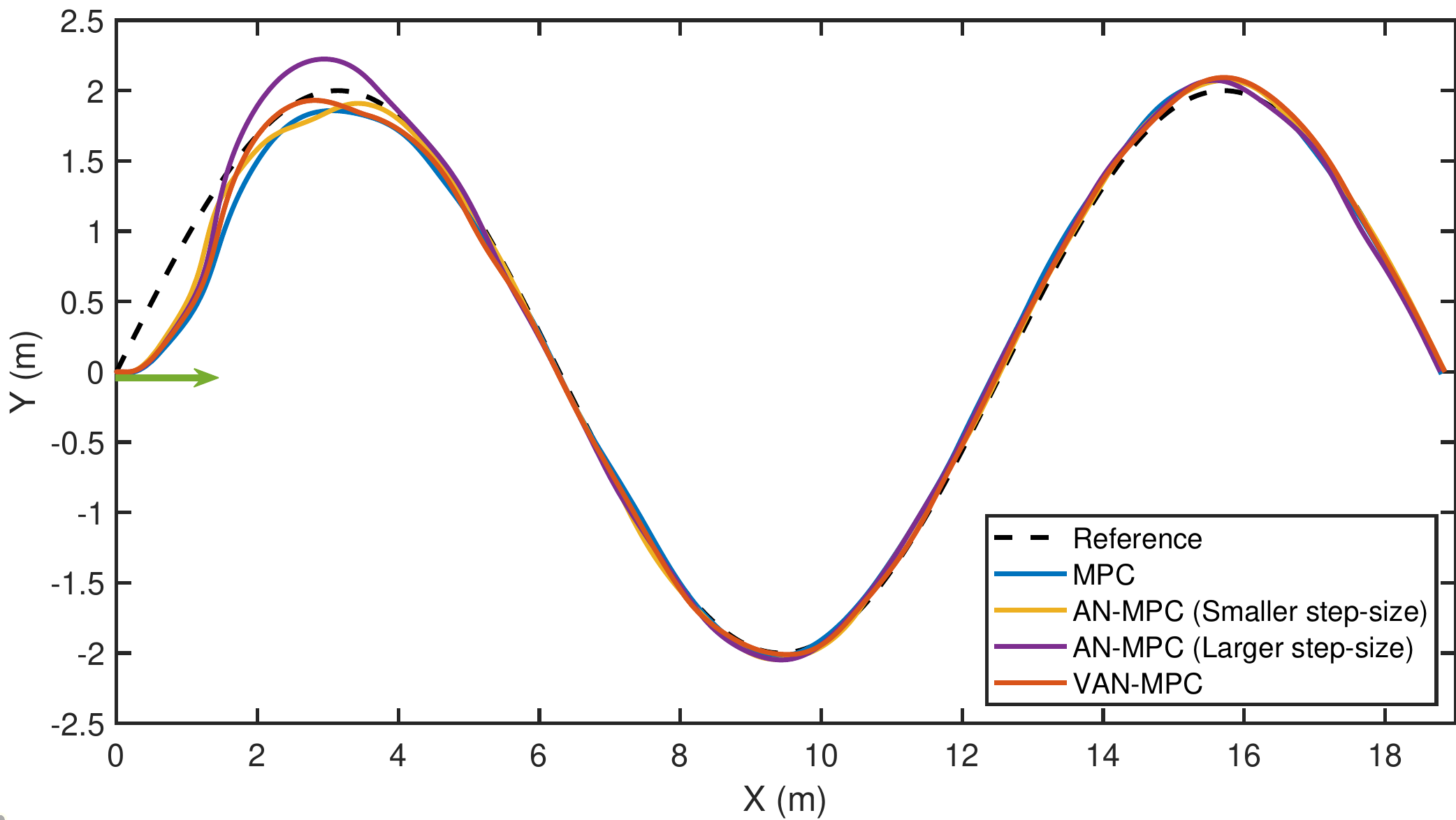}}
    \hspace{0in}    
\subfigure[Tracking distance]
{
    \label{fig13:subfig:b} 
    \includegraphics[width=3cm]{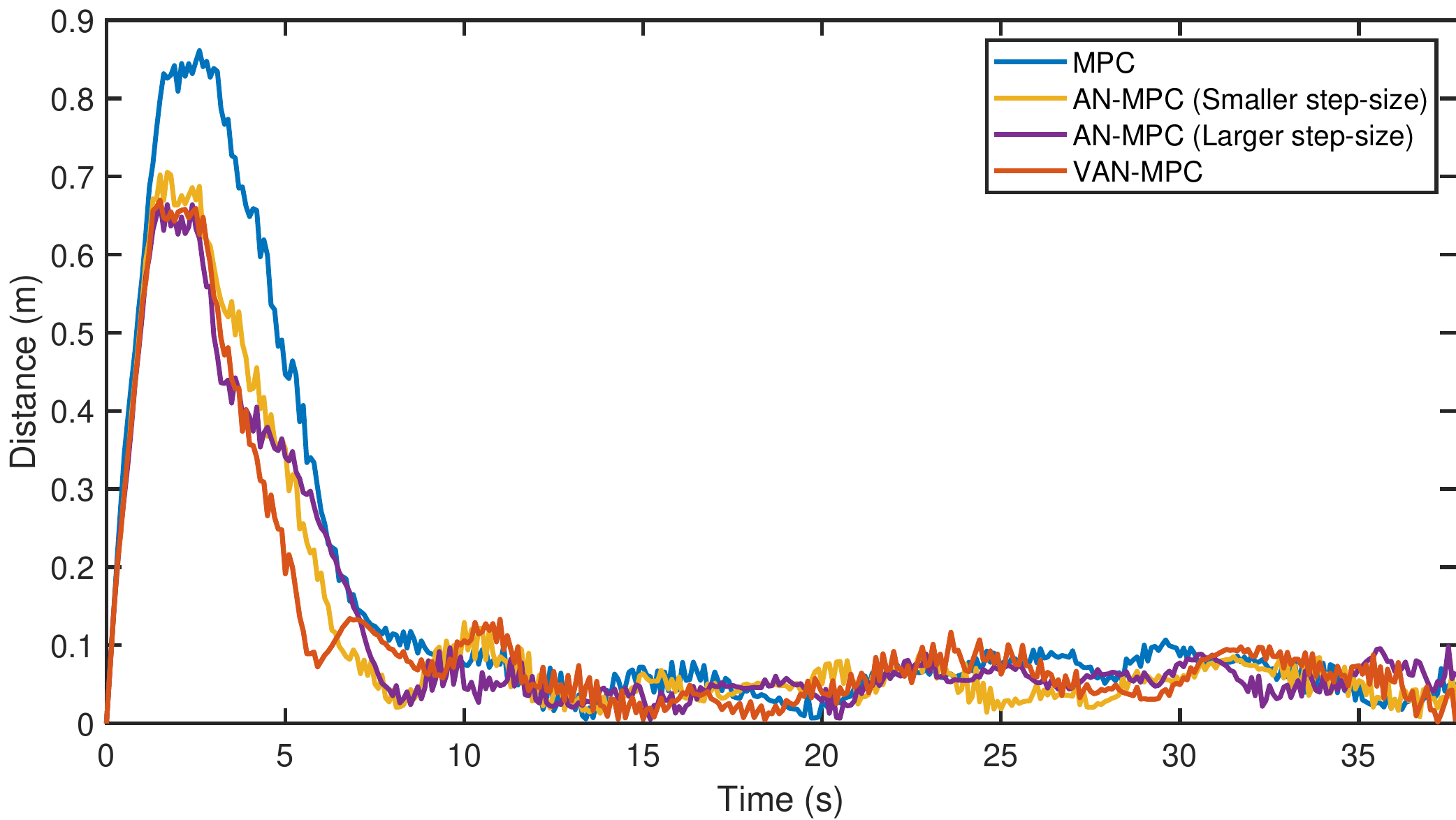}}
    \hspace{0in}
\caption{Experiment on flat tiled floor with no uncertainties.}
\label{fig13}
\end{figure}

\begin{figure}[t]
\setlength{\abovecaptionskip}{-2pt}
\centering
\subfigure[Real Path]
{
    \label{fig14:subfig:a} 
    \includegraphics[width=3cm]{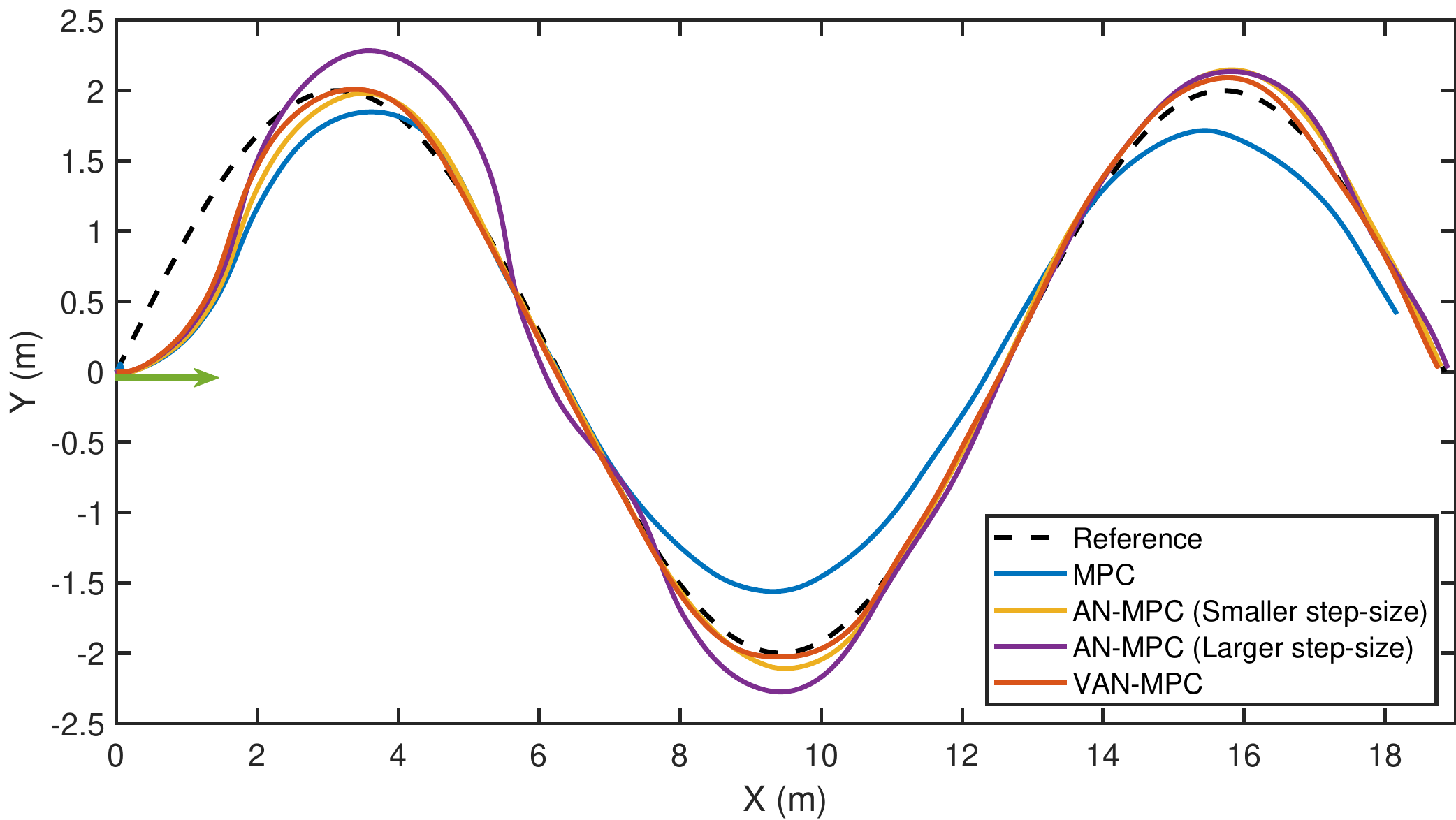}}
    \hspace{0in}    
\subfigure[Tracking distance]
{
    \label{fig14:subfig:b} 
    \includegraphics[width=3cm]{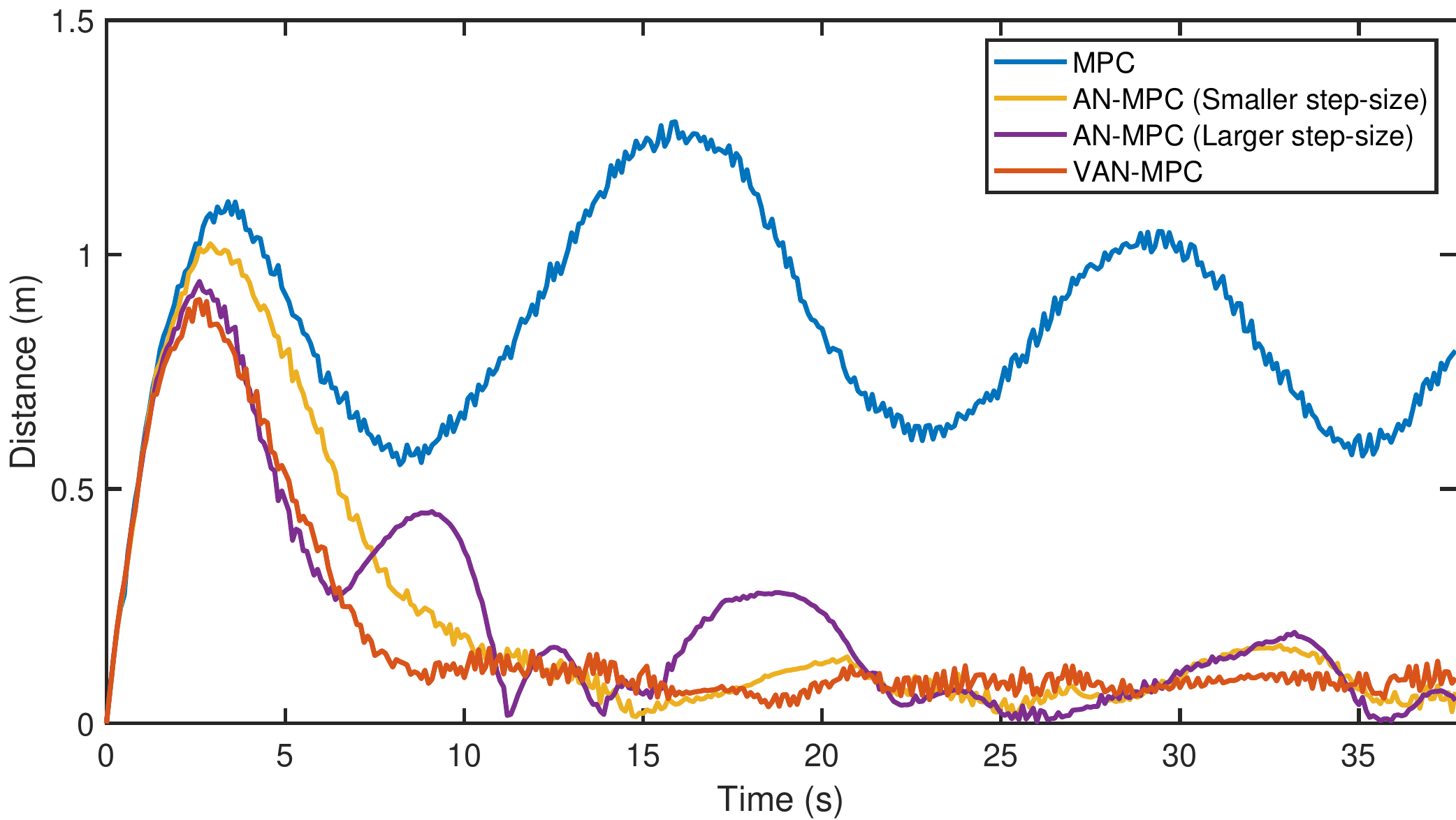}}
    \hspace{0in}
\caption{Experiment on flat tiled floor with artificial uncertainties $\boldsymbol{\Delta u}=\begin{bmatrix} 0.2v & 0 \end{bmatrix}$.}
\label{fig14}
\end{figure}

\begin{figure}[t]
\setlength{\abovecaptionskip}{-2pt}
\centering
\subfigure[Real Path]
{
    \label{fig15:subfig:a} 
    \includegraphics[width=3cm]{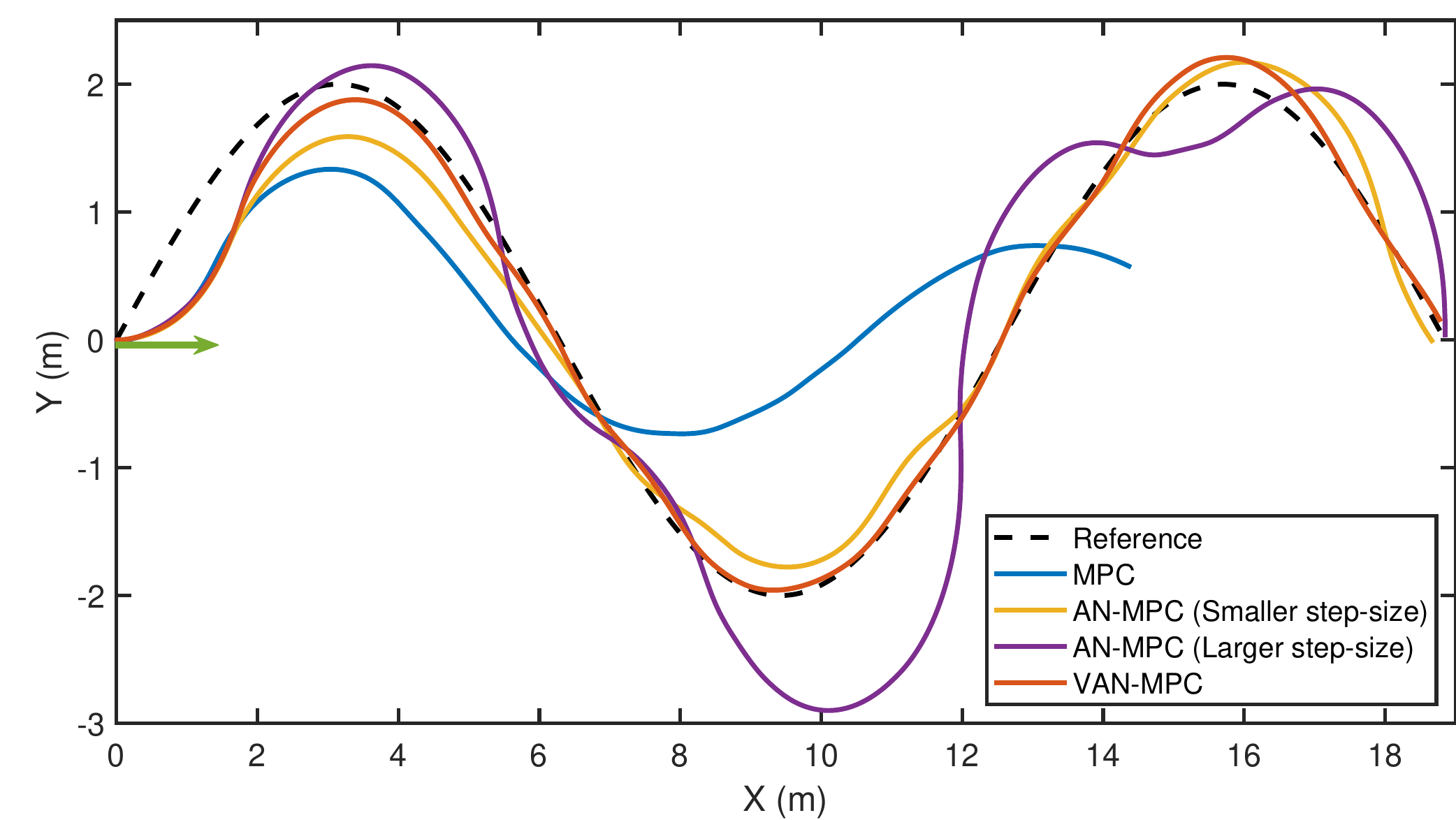}}
    \hspace{0in}    
\subfigure[Tracking distance]
{
    \label{fig15:subfig:b} 
    \includegraphics[width=3cm]{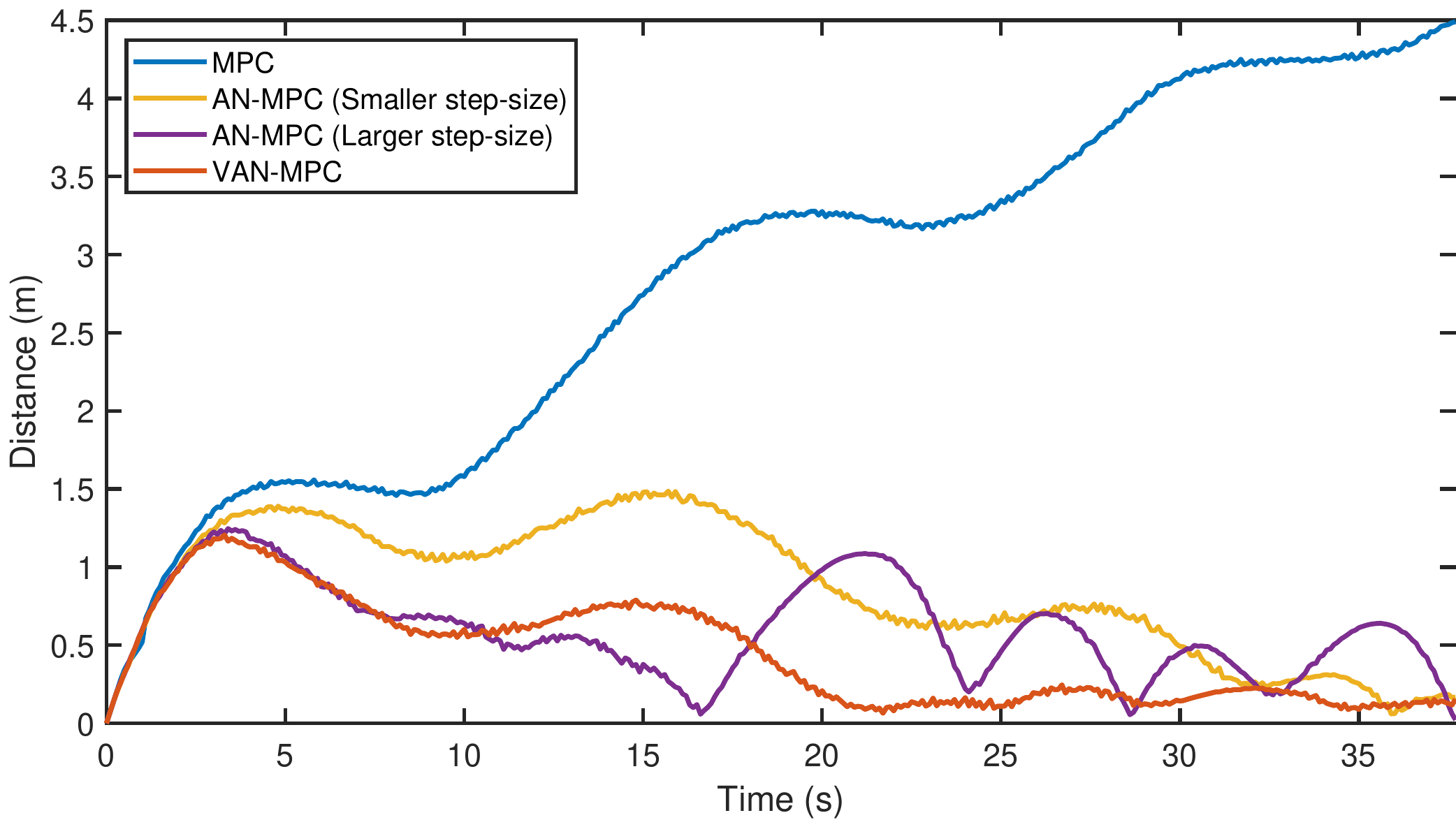}}
    \hspace{0in}
\caption{Experiment on flat tiled floor with artificial uncertainties $\boldsymbol{\Delta u}=\begin{bmatrix} 0.4v & 0 \end{bmatrix}$.}
\label{fig15}
\end{figure}

\begin{figure}[t]
\setlength{\abovecaptionskip}{-2pt}
\centering
\subfigure[$\Delta v=0.2v$]
{
    \label{fig16:subfig:a} 
    \includegraphics[width=3cm]{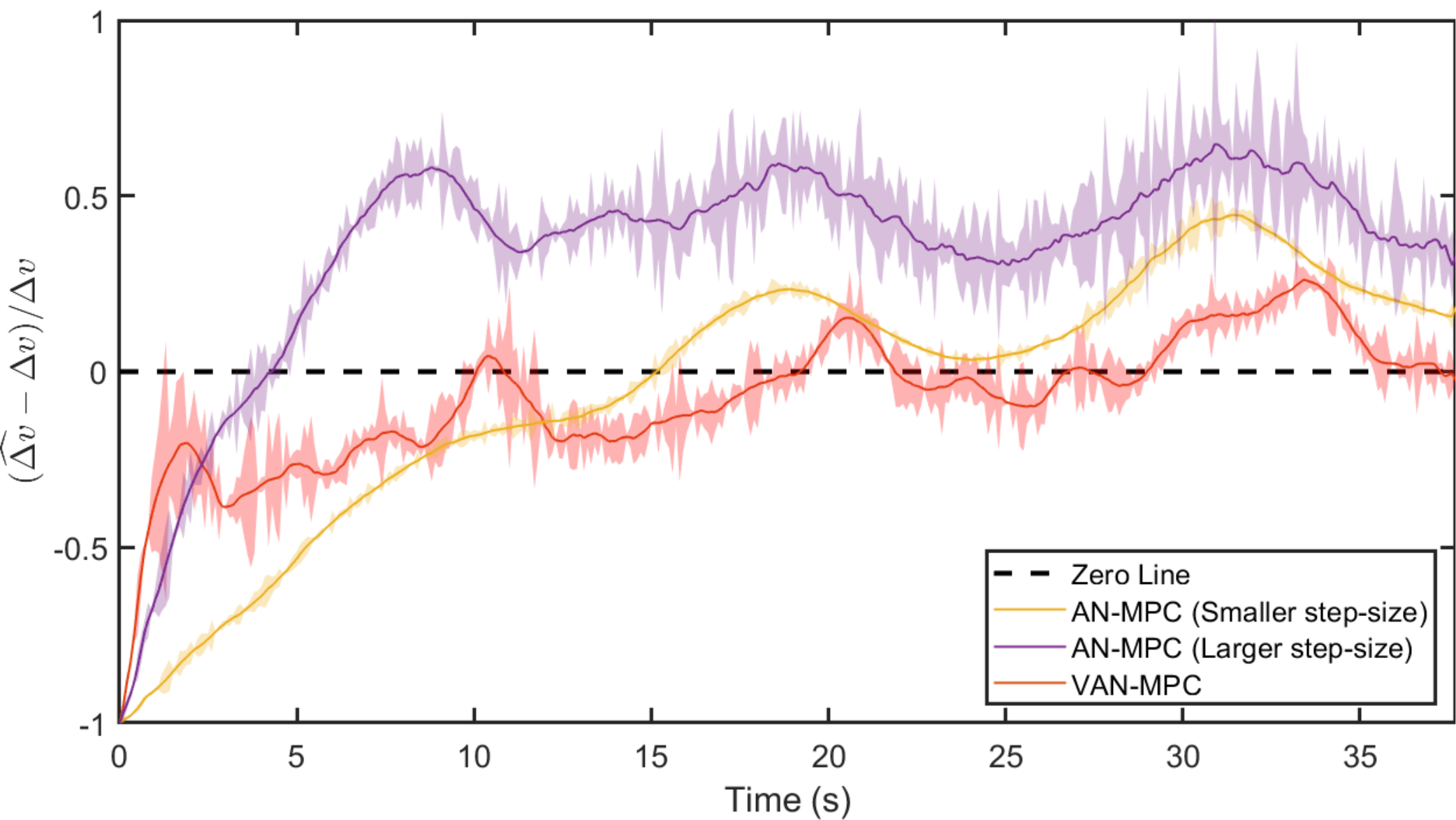}}
\subfigure[$\Delta v=0.4v$]
{
    \label{fig16:subfig:b} 
    \includegraphics[width=3cm]{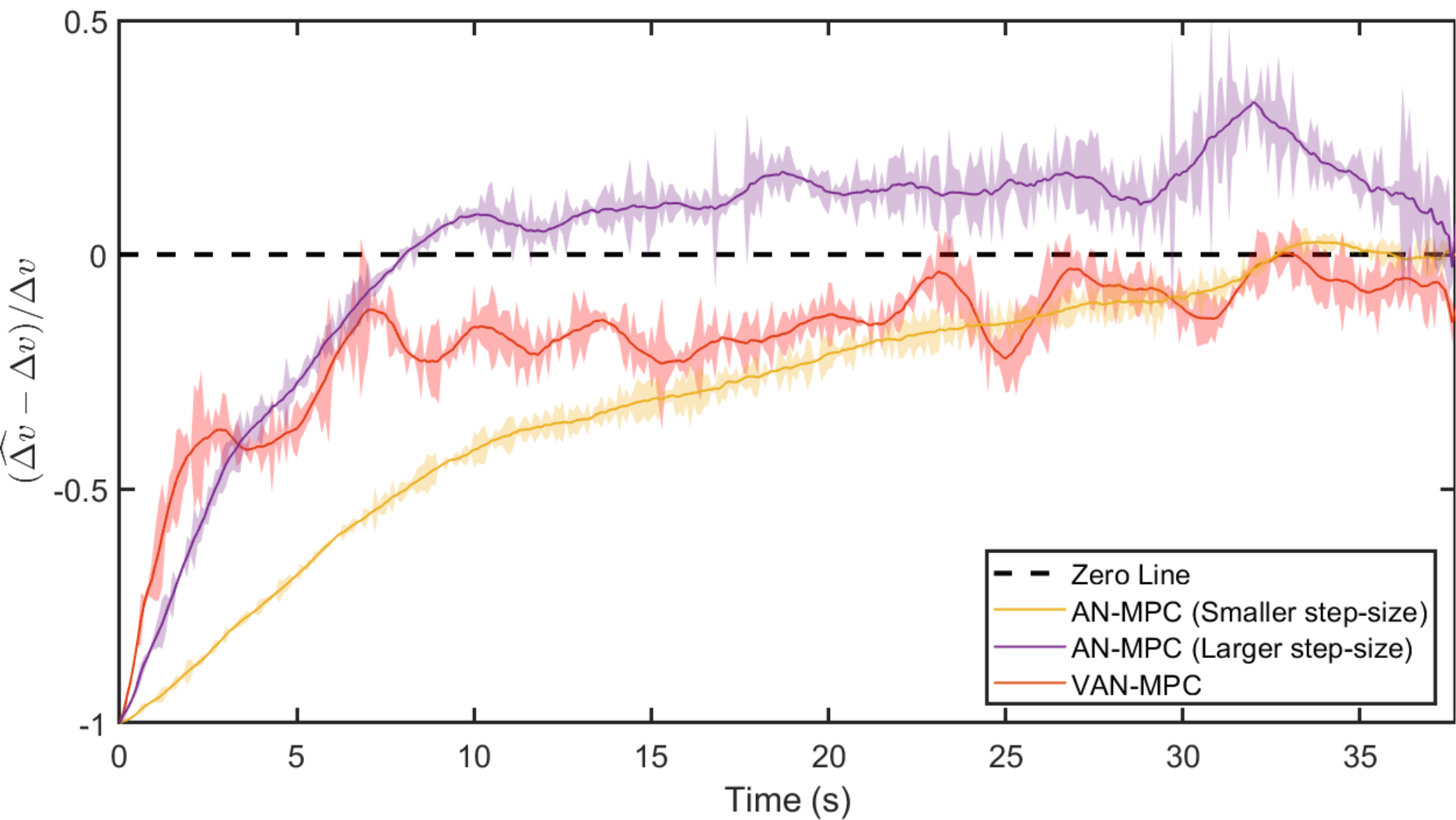}}
\caption{Relative error of the estimated uncertainties which indicate the convergence process of the estimated uncertainties.}
\label{fig16}
\end{figure}
\subsection{Flat Terrain with Artificial Uncertainties}

In reality, the uncertainties on certain terrain are unknown, and can vary with inputs, states, and the terrain's ups and downs, making it impossible to conduct quantitative analyses and determine whether our estimated uncertainties converge to the true value. Thus we hope to create artificial uncertainties and incorporate them into experiments to observe the convergence process. Since terrain Fig.~\ref{fig11:subfig:a} is highly unique and should theoretically be devoid of uncertainties, we apply artificial uncertainties to this place. The possible way is to insert a module between the instruction planner and the bottom controllers like Fig.~\ref{fig12}. In this experiment, we employ $\boldsymbol{\Delta u}=\begin{bmatrix}
    \xi\cdot v & 0
\end{bmatrix}$ as the artificial uncertainties that will vary with the command. And we hope to examine the convergence process of the estimated uncertainties.

The sine-wave trajectory depicted below is selected as the reference trajectory, where $t\in[0, 12\pi]$. Results of the experiments are shown in Fig.~\ref{fig13}-\ref{fig16}, where green array represents the initial position and direction. To better compare the control effects, the following indicators were chosen. (1) Rise time $t_r$ when the distance falls below 0.2 m for the first time. (2) Mean value of the distance $d_{m}$. (3) Mean distance after rising time $d_{mr}$. (4) The value of the first peak in the distance curve $d_{fp}$, which can indicate the learning speed of the uncertainties in the initial stage. (5) and (6) Root mean square error (RMSE) values of the estimated uncertainties' error boundaries $e_{rmsv}$ and $e_{rmsq}$ for $\widehat{\Delta v}$ and $\widehat{\Delta q_r}$, respectively. (7) Uncertainty rising time $t_{re}$ which is the time when the relative error of uncertainties first becomes zero. (8) RMSE of uncertainties' relative error $e_{rmser}$. Results are shown in Table.~\ref{table1}.

\begin{equation}
\begin{cases}
{X_{ref}}&=\;\;0.5t\\
{Y_{ref}}&=\;\;2\sin{(0.25t)}\\
{\phi_{ref}} &=\;\; \arctan{\cos(0.25t)}
\label{eq35}
\end{cases}
\end{equation}

\newcommand{\tabincell}[2]{\begin{tabular}{@{}#1@{}}#2\end{tabular}} 
\begin{table}[b]
\setlength{\abovecaptionskip}{-5pt}
\vspace{-0.5cm}
\begin{center}
\linespread{1.9}
\caption{Trajectory Tracking on Flat Tiled Floor}
\resizebox{1.0\columnwidth}{!}{
\begin{tabular}{cccccccccc}
\toprule
\multirow{2}*{\textbf{Uncertainty}}&\multirow{2}*{\tabincell{c}{\textbf{Instruction}\\\textbf{Planner}}} &  \multicolumn{8}{c}{\textbf{Indicators}}\\
\cmidrule(lr){3-10}
&&\textbf{\textit{$\boldsymbol{t_r}$}(s)}&\textbf{\textit{$\boldsymbol{d_m}$}(m)} & \textbf{\textit{$\boldsymbol{d_{mr}}$}(m)}&\textbf{\textit{$\boldsymbol{d_{fp}}$}(m)}
&\textbf{\textit{$\boldsymbol{e_{rmsv}}$}(m)}&\textbf{\textit{$\boldsymbol{e_{rmsq}}$}(m)} &\textbf{\textit{$\boldsymbol{t_{re}}$}(s)}&\textbf{\textit{$\boldsymbol{e_{rmser}}$}(m)}\\
\midrule
\multirow{4}*{\tabincell{c}{\textbf{With No}\\\textbf{Uncertainty}}}
&MPC & 6.5 & 0.1515 & 0.0645 & 0.8614 & None & None & None & None \\
&AN-MPC (Small) & 5.9 & 0.1186 & \textbf{0.0548} & 0.7055 & 0.0061 & 9.10E-5 & None & None \\
&AN-MPC (Large) & 6.5 & 0.1179 & 0.0555 & \textbf{0.6671} & 0.0240 & 1.02E-4 & None & None \\
&VAN-MPC & \textbf{4.9} & \textbf{0.1154} & 0.0640 & 0.6699 & \textbf{0.0048} & \textbf{5.67E-5} & None & None \\
\cmidrule(lr){2-10}
\multirow{4}*{$\boldsymbol{\Delta v=0.2v}$}
&MPC & None & 0.8425 & None & 1.113 & None & None & None & None \\
&AN-MPC (Small) & 9.5 & 0.2215 & \textbf{0.0914} & 1.024 & \textbf{0.0070} & 8.66E-5 & 15.1 & 0.3638 \\
&AN-MPC (Large) & 10.7 & 0.2228 & 0.1107 & 0.9430 & 0.0275 & 1.90E-4 & 3.3 & 0.4749 \\
&VAN-MPC & \textbf{7.2} & \textbf{0.1825} & 0.0932 & \textbf{0.9043} & 0.0118 & \textbf{6.53E-5} & \textbf{1.3} & \textbf{0.2166} \\
\cmidrule(lr){2-10}
\multirow{4}*{$\boldsymbol{\Delta v=0.4v}$}
&MPC & None & 2.8338 & None & 1.553 & None & None & None & None \\
&AN-MPC (Small) & 35.4 & 0.859049 & \textbf{0.1461} & 1.39 & \textbf{0.0191} & 4.05E-4 & 31.7 & 0.4119\\
&AN-MPC (Large) & \textbf{16.1} & 0.5986 & 0.5234 & 1.247 & 0.0488 & 7.24E-4 & 7.9 & 0.2738 \\
&VAN-MPC & 19.9 & \textbf{0.4474} & 0.1533 & \textbf{1.206} & 0.0271 & \textbf{1.50E-4} & \textbf{6.8} & \textbf{0.2489} \\
\bottomrule
\end{tabular}}
\label{table1}
\end{center}
\end{table}

When there is no uncertainties, each of the four frameworks can effectively work according to Fig.~\ref{fig13}. And it can be seen that the VAN-MPC and AN-MPC can slow the increase in distance and reduce the error faster. When artificial uncertainties are inserted, according to Fig.~\ref{fig14}-\ref{fig16} and Table.~\ref{table1}, the effects of the four tracking frameworks vary widely. MPC cannot work in such situations with the largest distances while VAN-MPC has the best tracking effect with small $t_r$, $d_m$, $d_{mr}$ and $d_{fp}$. VAN-MPC and can reduce tracking error and maintain stability rapidly which is due to the fact that it initially has a large step-size, but the step-size will progressively decrease to maintain stability. According to Fig.~\ref{fig16}, it's obvious that VAN-MPC can converge quickly to the zero line at the beginning with the smallest $t_{re}$, and then maintain stability with the smallest $e_{rmser}$.

\begin{figure}[t]
\setlength{\abovecaptionskip}{-2pt}
\centering
\subfigure[Real Path]
{
    \label{fig17:subfig:a} 
    \includegraphics[width=3cm]{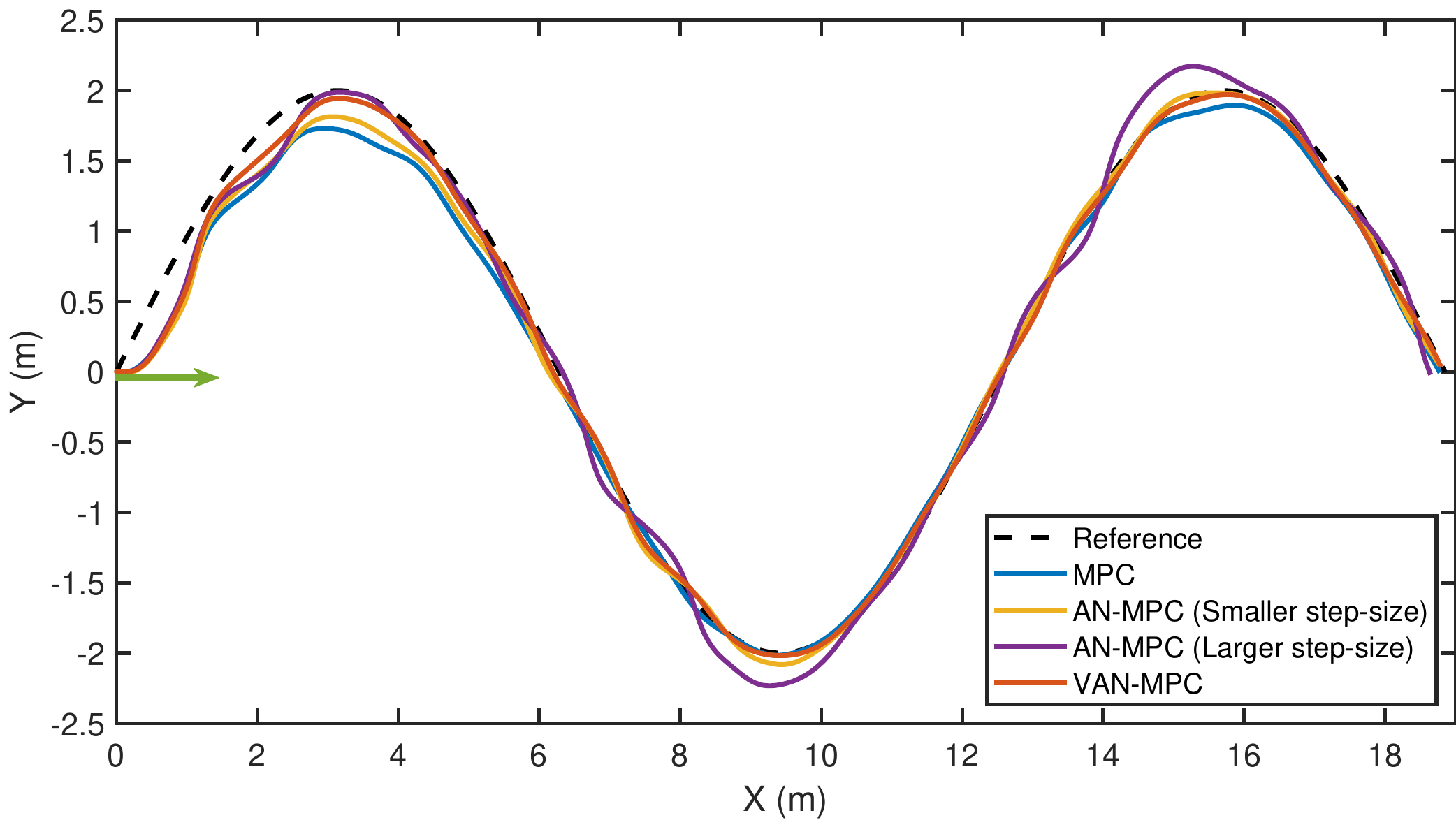}}
    \hspace{0in}    
\subfigure[Tracking distance]
{
    \label{fig17:subfig:b} 
    \includegraphics[width=3cm]{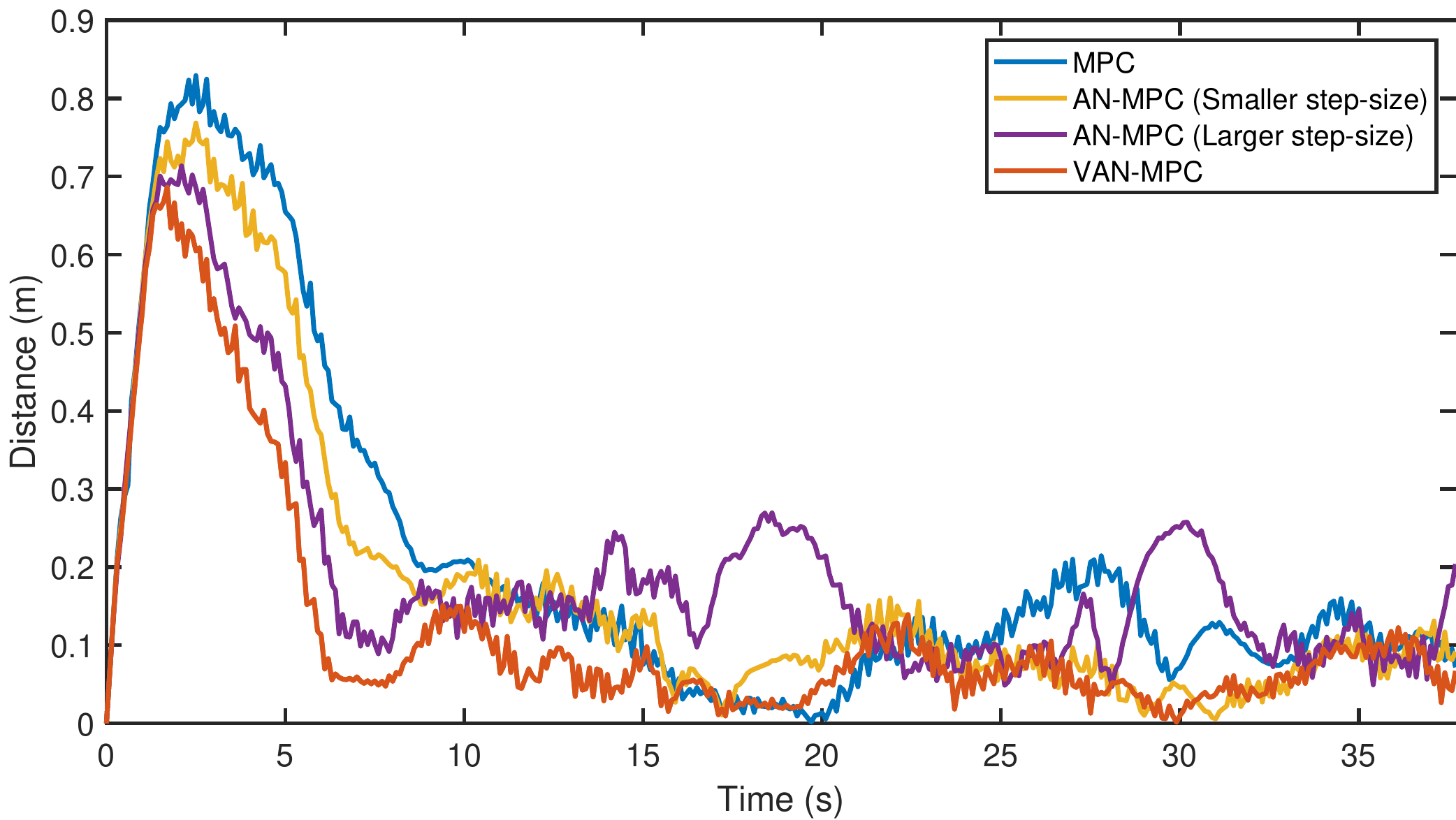}}
    \hspace{0in}
\caption{Experiment on rubber ground.}
\label{fig17}
\end{figure}

\begin{figure}[t]
\setlength{\abovecaptionskip}{-2pt}
\centering
\subfigure[Real Path]
{
    \label{fig18:subfig:a} 
    \includegraphics[width=3cm]{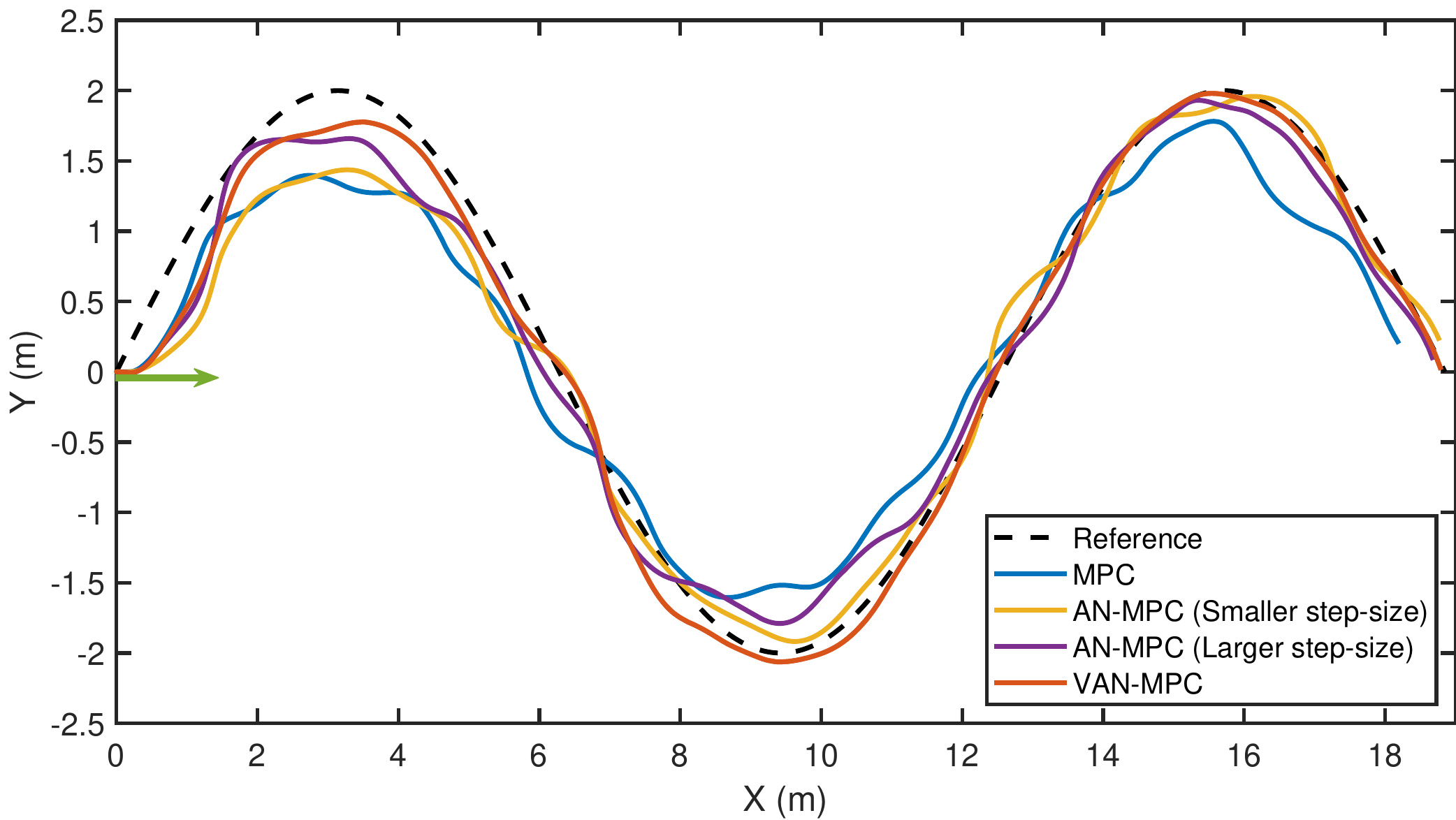}}
    \hspace{0in}    
\subfigure[Tracking distance]
{
    \label{fig18:subfig:b} 
    \includegraphics[width=3cm]{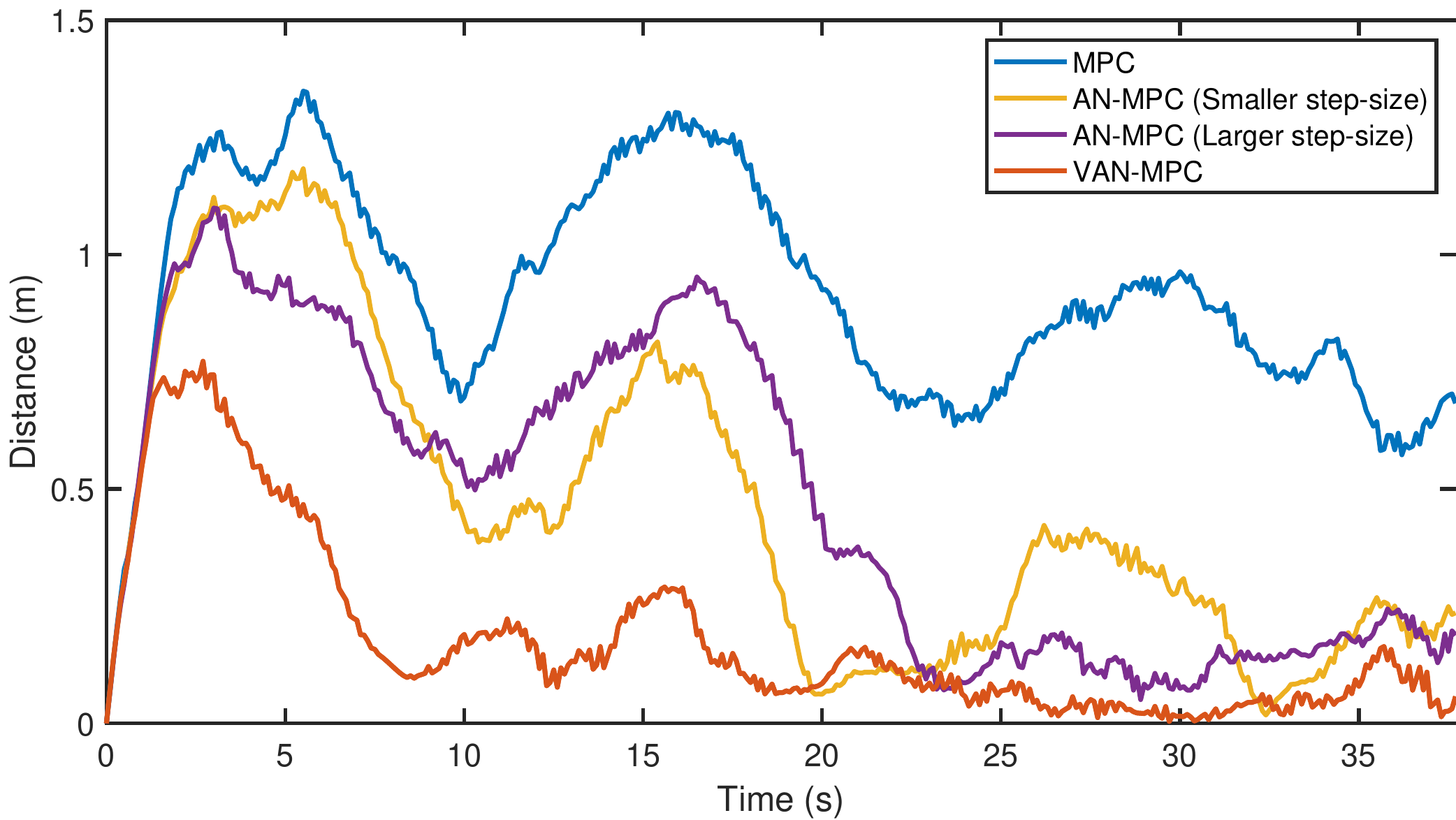}}
    \hspace{0in}
\caption{Experiment on ground with hollow tiles.}
\label{fig18}
\end{figure}

\begin{figure}[t]
\setlength{\abovecaptionskip}{-2pt}
\centering
\subfigure[Real Path]
{
    \label{fig19:subfig:a} 
    \includegraphics[width=3cm]{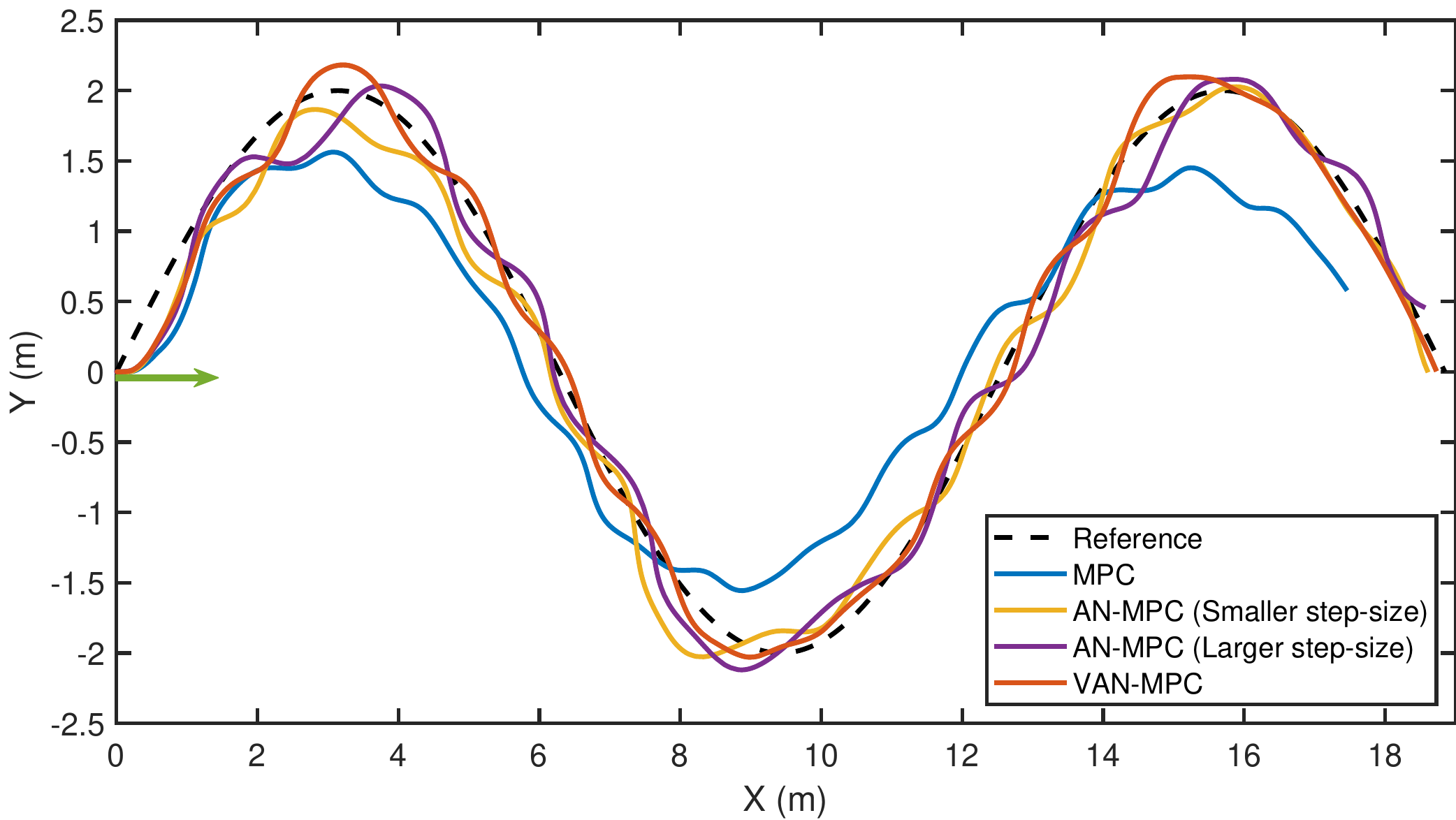}}
    \hspace{0in}    
\subfigure[Tracking distance]
{
    \label{fig19:subfig:b} 
    \includegraphics[width=3cm]{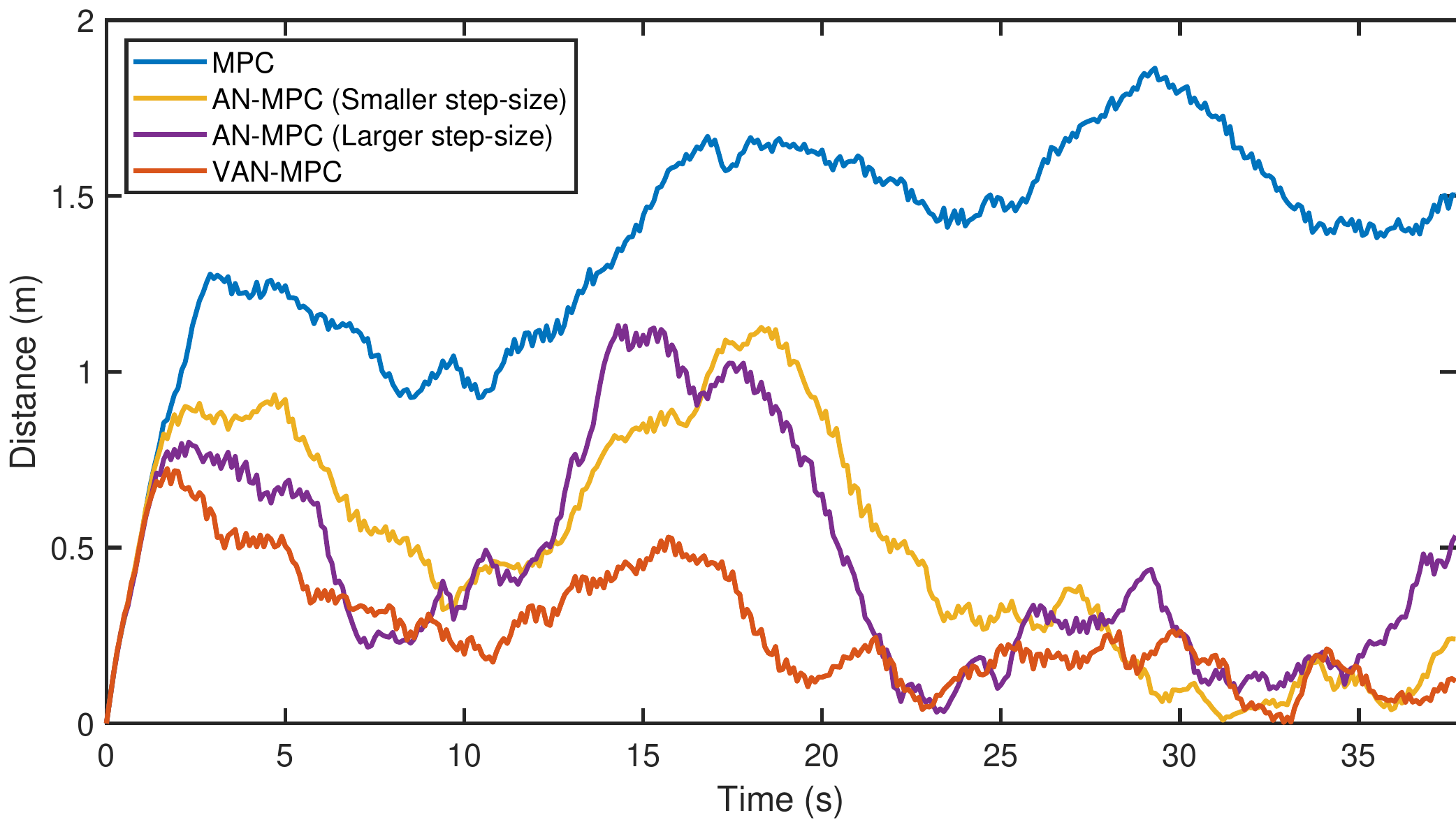}}
    \hspace{0in}
\caption{Experiment on grass.}
\label{fig19}
\end{figure}

\begin{figure}[t]
\setlength{\abovecaptionskip}{-2pt}
\centering
\subfigure[Real Path]
{
    \label{fig20:subfig:a} 
    \includegraphics[width=3cm]{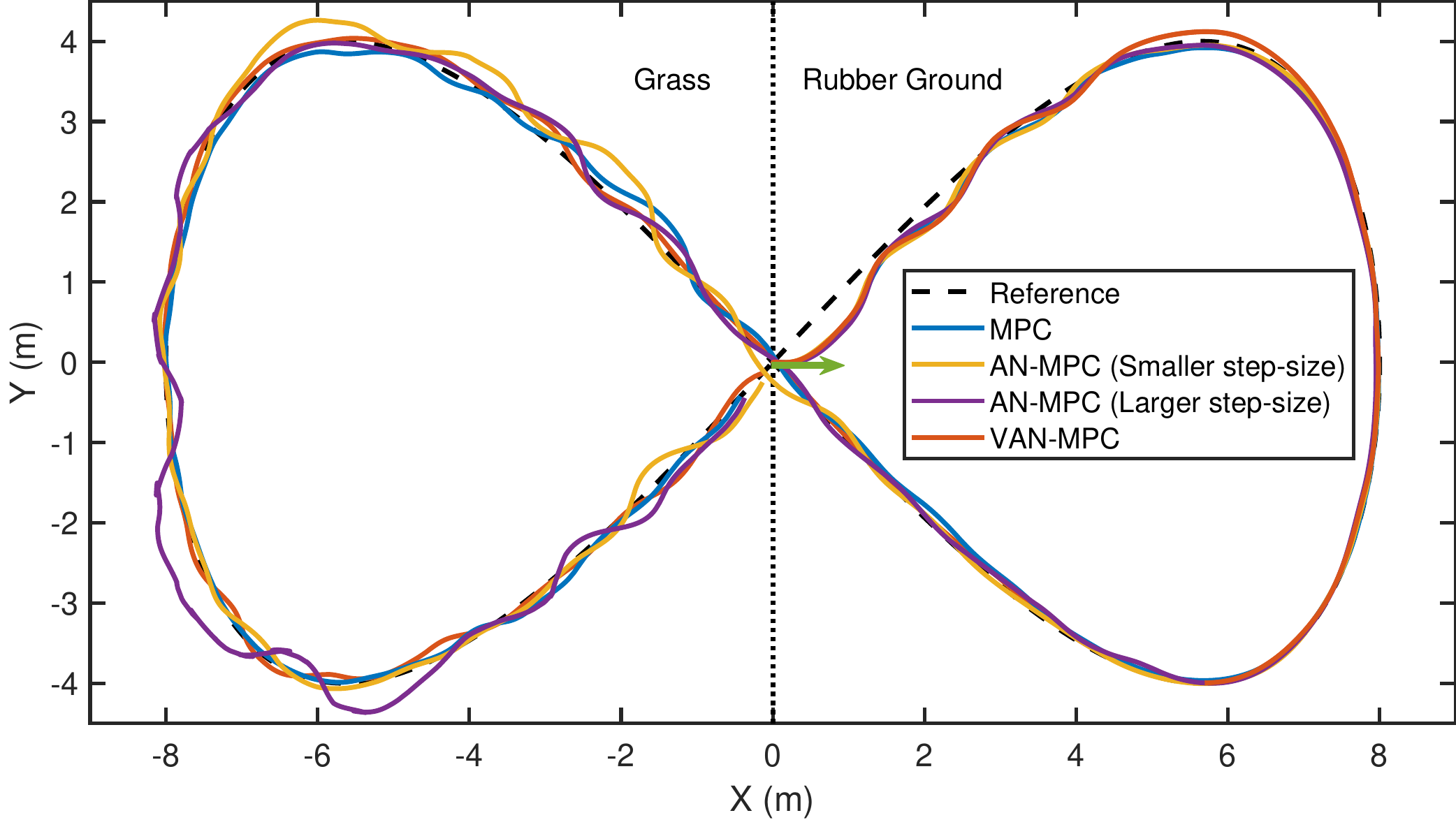}}
    \hspace{0in}    
\subfigure[Tracking distance]
{
    \label{fig20:subfig:b} 
    \includegraphics[width=3cm]{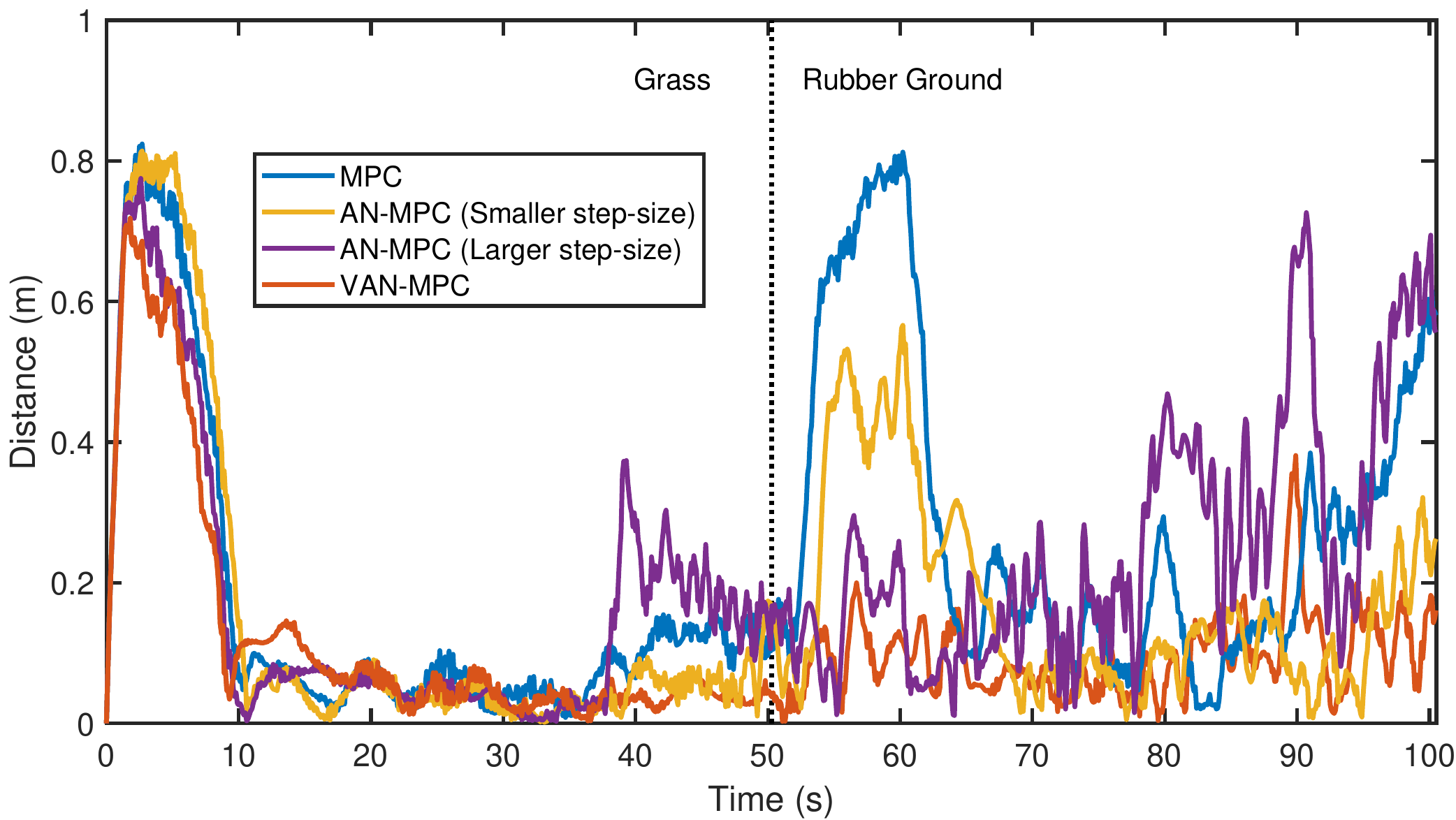}}
    \hspace{0in}
\caption{Experiment on varied terrain}
\label{fig20}
\end{figure}
\subsection{Experiments on Multiple Terrains}
In this subsection, we carry out trajectory tracking experiments on three terrains in Fig.~\ref{fig11:subfig:b}-\ref{fig11:subfig:d}. Due to the absence of the uncertainties' true value, we consider the entire system to be a black box and evaluate the algorithms based on the distance curves. Results of the experiments are shown in Fig.~\ref{fig17}-\ref{fig19} with the reference trajectory in \eqref{eq35}. Results of indicators can be seen in Table.~\ref{table2}. According to the results, VAN-MPC has much better control effect over the others on multiple terrains. On rubber terrain, the uncertainties are relatively small, and MPC can help progressively reduce the distance. But the uncertainties will change with the desired velocity, causing the distance curve to fluctuate accordingly. However, MPC is unable to manage situations with large uncertainties like the ground with hollow tiles and grass. In contrast, other planners with estimated uncertainties can decrease the distance more quickly, particularly VAN-MPC, which can stop the rise in distance the quickest with the smallest $d_fp$ on all three terrains. As a result of its large step-size in the initial stage, VAN-MPC also has a small $t_r$. When the distance or uncompensated uncertainties increase, the step-size of VAN-MPC may increase quickly and then decreases to maintain stability and have the smallest $d_{mr}$ and $d_m$. This makes the trajectory tracking control framework VANMHH equipped with VAN-MPC more robust and efficient, even for rugged terrains. As for the AN-MPC (large step-size), due to its constant large step-size, the estimated uncertainties fluctuate violently. Although the estimated uncertainties of the AN-MPC (small step-size) are comparatively stable and fluctuate less with small $e_{rmsv}$ and $e_{rmsq}$, they changes so slowly that they cannot adapt well to changes in environmental uncertainties. The two aforementioned factors cause the distance curves of the two AN-MPC algorithms to vary. As a conclusion, VANMHH equipped with VAN-MPC performs best on a variety of terrains due to its superior stability, robustness, and response speed.

\begin{table}[b]
\setlength{\abovecaptionskip}{-3pt}
\begin{center}
\vspace{-0.5cm}
\linespread{1.9}
\caption{Trajectory Tracking on Multiple Terrains}
\resizebox{.9\columnwidth}{!}{
\begin{tabular}{cccccccc}
\toprule
\multirow{2}*{\textbf{Uncertainty}}&\multirow{2}*{\tabincell{c}{\textbf{Instruction}\\\textbf{Planner}}} &  \multicolumn{6}{c}{\textbf{Indicators}}\\
\cmidrule(lr){3-8}
&&\textbf{\textit{$\boldsymbol{t_r}$}(s)}&\textbf{\textit{$\boldsymbol{d_m}$}(m)} & \textbf{\textit{$\boldsymbol{d_{mr}}$}(m)}&\textbf{\textit{$\boldsymbol{d_{fp}}$}(m)}
&\textbf{\textit{$\boldsymbol{e_{rmsv}}$}(m)}&\textbf{\textit{$\boldsymbol{e_{rmsq}}$}(m)}\\
\midrule
\multirow{4}*{\tabincell{c}{\textbf{Rubber}\\\textbf{Ground}}}
&MPC            & 8.9 & 0.2093 & 0.1092 & 0.8293 & None & None \\
&AN-MPC (Small) & 8.0 & 0.1783 & 0.0941 & 0.7686 & 0.0073 & 3.88E-4 \\
&AN-MPC (Large) & 6.3 & 0.1985 & 0.1424 & 0.7137 & 0.0362 & 7.28E-4 \\
&VAN-MPC        & \textbf{5.6} & \textbf{0.1228} & \textbf{0.0662} & \textbf{0.6864} & \textbf{0.0067} & \textbf{1.20E-4}\\
\cmidrule(lr){2-8}
\multirow{4}*{\tabincell{c}{\textbf{Ground with}\\\textbf{Hollow Tiles}}}
&MPC & None & 0.9210 & None & 1.3490 & None & None \\
&AN-MPC (Small) & 19.3 & 0.4615 & 0.2053 & 1.1830 & 0.0110 & 6.09E-4 \\
&AN-MPC (Large) & 22.4 & 0.4767 & 0.1425 & 1.0990 & 0.0410 & 7.33E-4 \\
&VAN-MPC & \textbf{7.1} & \textbf{0.1794} & \textbf{0.1025} & \textbf{0.7726} & \textbf{0.0081} & \textbf{2.70E-4}\\
\cmidrule(lr){2-8}
\multirow{4}*{\textbf{Grass}}
&MPC & None & 1.3654 & None & 1.278 & None & None \\
&AN-MPC (Small) & 28.4 & 0.4981 & \textbf{0.1028} & 0.9352 & \textbf{0.0118} & 0.0014\\
&AN-MPC (Large) & 21.7 & 0.4605 & 0.2251 & 0.7998 & 0.0755 & 0.0020 \\
&VAN-MPC & \textbf{10.0} & \textbf{0.2701} & 0.2151 & \textbf{0.7245} & 0.0273 & \textbf{0.0013}\\
\bottomrule
\end{tabular}}
\label{table2}
\end{center}
\end{table}

\subsection{Experiments on Varied Terrain}

Experiments on the varied terrain in Fig.~\ref{fig11:subfig:e} is conducted in this subsection. The reference trajectory is a Lemniscate of Gerono, which is shown in \eqref{eq36} where $t\in[0, 32\pi]$. The robot will transit from rubber ground which locates on the right to the grass. Experiment results are shown in Fig.~\ref{fig20}. According to Fig.~\ref{fig20}, VAN-MPC has the best tracking performance, and when the terrain changes, it can also very effectively eliminate the rise in distance so as to keep a stable and effective following effect. Moreover, it is obvious that VAN-MPC has the least global tracking error and the smallest tracking error on both two terrains. In conclusion, VANMHH equipped with VAN-MPC is still capable of maintaining an effective trajectory tracking effect on varied terrain.

\begin{equation}
\begin{cases}
{X_{ref}}&=8\sin{(t/16)}\\
{Y_{ref}}&=8\sin{(t/16)}\cos{(t/16)}\\
{\phi_{ref}} &= \text{atan2}\left(\cos{(t/8)}/\cos{(t/16)} S_{ign}, \;\;  S_{ign}\right)\\
S_{ign} &= \text{sgn}\left(\cos{(t/16)}\right)
\label{eq36}
\end{cases}
\end{equation}

\section{CONCLUSIONS}

Five terrains are selected, and real world experiments are carried out to verify the control effect. 

In this paper, we develop an efficient control framework VANMHH for the multi-terrain trajectory tracking problem of the spherical robot. This new framework is developed by enhancing our previous MHH framework, and offers multi-terrain capabilities by replacing the existing instruction planner MPC with VAN-MPC. All the uncertainties in both kinematics and dynamics are rearranged and estimated by a modified RBFNN, which is then sent to the VAN-MPC planner. Then, VAN-MPC with an updated model will solve the new optimal problem in order to determine the ideal command, which will be sent to HSMC and HTSMC. As a result, spherical robots can track trajectories on multiple terrains effectively with VANMHH. 

Then five terrains are selected, and real world experiments are carried out to verify the control effect. First, we apply different artificial uncertainties, and observe that the uncertainties can rapidly converge to the true value and the framework has a great tracking effect in the presence of various artificial uncertainties. Then, experiments on other terrains and varied terrain demonstrate that the proposed method is very effective and enables the spherical robot to achieve trajectory tracking on multiple terrains, including uneven terrains and unknown terrains. And this study can help spherical robots have broader applications. 

In the future, we will conduct an in-depth study on spherical robots from the perspective of task decision-making, so that spherical robots can truly execute tasks such as exploration and rescue in the wild, and make contributions to human beings.

\bibliographystyle{IEEEtranTIE}
\bibliography{main.bbl}\ 

\end{document}